\documentclass[journal]{IEEEtran}

\usepackage{times}
\usepackage{epsfig}
\usepackage{graphicx}
\usepackage{amsmath}
\usepackage{amssymb}
\usepackage{blindtext}
\usepackage[ruled,vlined,linesnumbered]{algorithm2e}
\usepackage{multirow}
\usepackage{booktabs}
\usepackage{subfigure}
\usepackage{cite}
\usepackage[pagebackref=true,breaklinks=true,letterpaper=true,colorlinks,bookmarks=false]{hyperref}
 
\usepackage{xspace}
\usepackage{makecell}
\usepackage[square,numbers]{natbib}
\usepackage{colortbl}

\makeatletter
\DeclareRobustCommand\onedot{\futurelet\@let@token\@onedot}
\def\@onedot{\ifx\@let@token.\else.\null\fi\xspace}

\def\eg{\emph{e.g}\onedot} 
\def\ie{\emph{i.e}\onedot}

\def\etal{\emph{et al}\onedot}
\makeatother

\definecolor{graycolor}{rgb}{0.95,0.95,0.95}

\SetKwInput{KwInput}{Input}                
\SetKwInput{KwOutput}{Output}

\ifCLASSINFOpdf
\else
\fi

\begin{document}
%
\title{Hyperspectral Image Super Resolution with Real Unaligned RGB Guidance}
%
%
%

\author{Zeqiang Lai,
        Ying Fu,~\IEEEmembership{Senior Member,~IEEE}, 
        and Jun Zhang
}
%
%

\markboth{Journal of \LaTeX\ Class Files,~Vol.~14, No.~8, August~2015}%
{Shell \MakeLowercase{\textit{et al.}}: Bare Demo of IEEEtran.cls for IEEE Journals}
%



\maketitle

\begin{abstract}
   Fusion-based hyperspectral image (HSI) super-resolution has become increasingly prevalent for its capability to integrate high-frequency spatial information from the paired high-resolution (HR) RGB reference image.
   However, most of the existing methods either heavily rely on the accurate alignment between low-resolution (LR) HSIs and RGB images, or can only deal with simulated unaligned RGB images generated by rigid geometric transformations, which weakens their effectiveness for real scenes.
   In this paper, we explore the fusion-based HSI super-resolution with real RGB reference images that have both rigid and non-rigid misalignments. To properly address the limitations of existing methods for unaligned reference images, we propose an HSI fusion network with heterogenous feature extractions, multi-stage feature alignments, and attentive feature fusion. Specifically, our network first transforms the input HSI and RGB images into two sets of multi-scale features with an HSI encoder and an RGB encoder, respectively. The features of RGB reference images are then processed by a multi-stage alignment module to explicitly align the features of RGB reference with the LR HSI. Finally, the aligned features of RGB reference are further adjusted by an adaptive attention module to focus more on discriminative regions before sending them to the fusion decoder to generate the reconstructed HR HSI. Additionally, we collect a real-world HSI fusion dataset, consisting of paired HSI and unaligned RGB reference, to support the evaluation of the proposed model for real scenes. 
   Extensive experiments are conducted on both simulated and our real-world datasets, and it shows that our method obtains a clear improvement over existing single-image and fusion-based super-resolution methods on quantitative assessment as well as visual comparison.
The code and dataset are publicly available at \url{https://zeqiang-lai.github.io/HSI-RefSR/}.
\end{abstract}


%
\IEEEpeerreviewmaketitle

\section{Introduction}
Hyperspectral imaging systems are designed to collect and process the abundant spectral information from across the electromagnetic spectrum. Different from conventional RGB cameras, spectral imaging systems divide the spectrum into much more bands than three, which provides higher spectral resolution. However, limited by the existing imaging techniques, higher spectral resolution often comes at the expense of lower spatial resolution. This might limit the applications of HSI in the fields of remote sensing \cite{blackburn2007hyperspectral,thenkabail2016hyperspectral,zhang2022artificial}, classification \cite{camps2005kernel,hu2015deep,rodarmel2002principal}, and etc \cite{bjorgan2015towards, pan2003face}.

\begin{figure}[t]
\centering
\subfigure[RGB Reference]{
\includegraphics[width=0.145\textwidth]{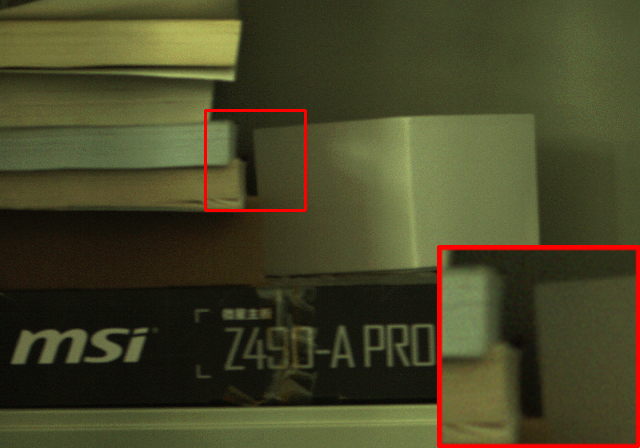}
\label{fig:f-rgb}
}
\subfigure[Ground Truth]{
\includegraphics[width=0.145\textwidth]{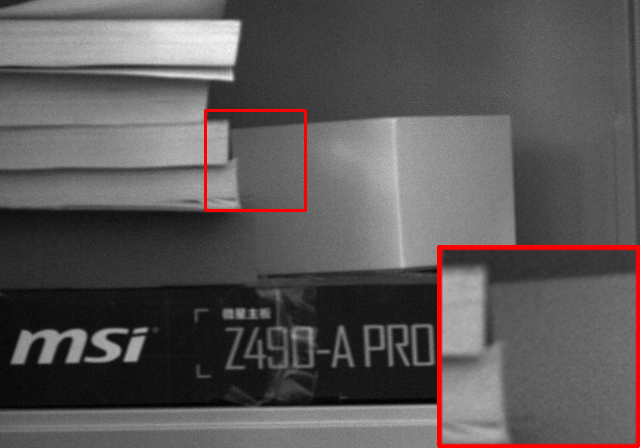}
}
\subfigure[LR HSI ($\times8$)]{
\includegraphics[width=0.145\textwidth]{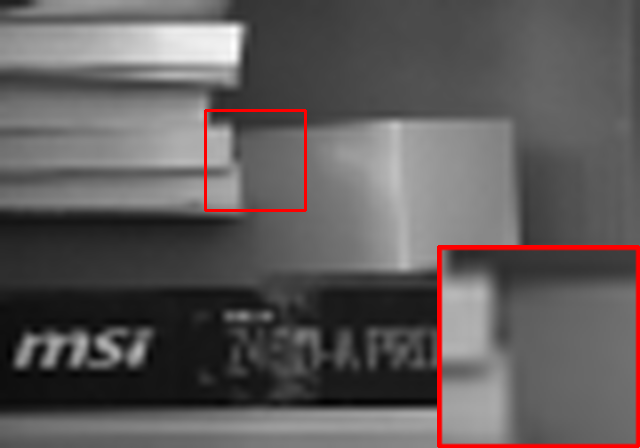}
}
\quad
\subfigure[SSPSR\cite{jiang2020learning}]{
\includegraphics[width=0.145\textwidth]{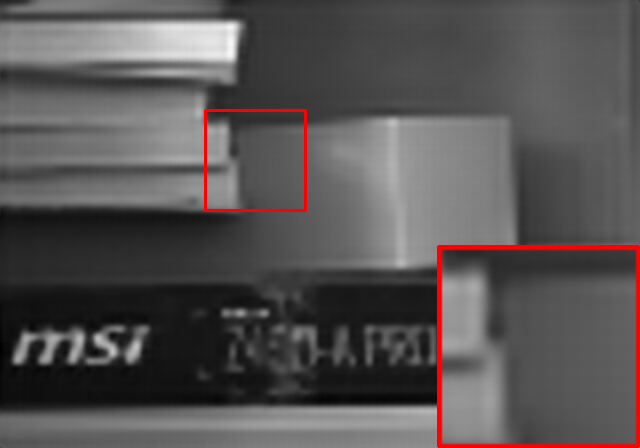}
}
\subfigure[Optimized\cite{fu2019hyperspectral}]{
\includegraphics[width=0.145\textwidth]{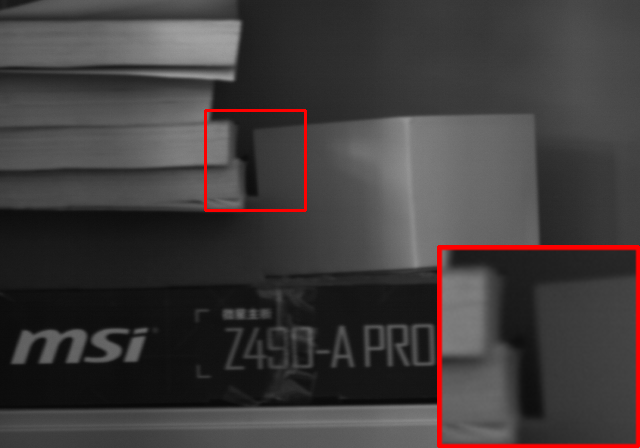}
}
\subfigure[Ours]{
\includegraphics[width=0.145\textwidth]{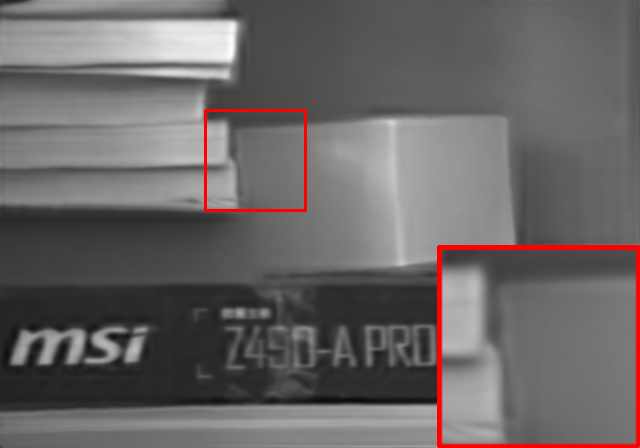}
}

\caption{Illustration of the misalignment between the real RGB reference image and LR HSI. It can be observed there is a tiny offset for the white box. For the comparison of different methods, our method produces the clearest result on the premise of proper alignment. The SISR method, \ie, SSPSR \cite{jiang2020learning}, obtains the aligned but more blurred result. The fusion-based method, \ie, Optimized \cite{fu2019hyperspectral}, processes directly on RGB reference without considering the misalignment.}

\label{fig:demo}
\end{figure}

With the aim of lifting the spatial resolution, most recent works \cite{li2020mixed,jiang2020learning,jiang2020learning} follow the paradigm of single image super-resolution (SISR) that upsamples the spatial resolution given the single LR HSI. These methods usually depend upon the powerful learning capability of different types of complex convolutional neural network (CNN) to reconstruct missing high-frequency details. For example, Li \etal \cite{li2020mixed} propose a mixed convolutional network (MCNet) by utilizing both 2D and 3D convolutions. Jiang \etal \cite{jiang2020learning} introduce SSPSR that explores the spatial and spectral prior with group convolution. Fu \etal \cite{fu2021bidirectional} extend the 3D-CNN with a bi-directional quasi-recurrent neural network to enhance the inter-spectral interactions. Though progress has been made, the performance of these approaches is still physically restricted by the deficient information provided by LR input, which hinders the further improvements, especially for large scaling factors. 

To overcome the limitation of SISR, alternative approaches \cite{chang2020weighted,qu2018unsupervised,fu2019hyperspectral} consider the HSI super-resolution in hybrid imaging systems, where an aligned HR RGB camera is used to complement the hyperspectral counterpart. 
With these systems, paired aligned data can be obtained and various optimization-based \cite{chang2020weighted, akhtar2014sparse} and CNN-based methods \cite{dian2018deep,fu2019hyperspectral} are proposed to transfer the high-frequency details from HR RGB reference image for the reconstruction of HR HSI from the captured LR HSI. These methods usually perform better than SISR approaches, but heavily rely on the complex imaging system and careful calibration to ensure precise alignment, which weakens its effectiveness for practical applications. To alleviate the strong assumption of existing fusion-based approaches, some recent works \cite{fu2020simultaneous,nie2020unsupervised, qu2021unsupervised, zheng2021nonregsrnet} begin to take into account the misalignment of RGB reference images, \eg, Fu \etal \cite{fu2020simultaneous} propose an alternating direction method of multipliers (ADMM)-based method for solving HSI super-resolution with rigid geometric misaligned RGB reference, Qu \etal \cite{qu2021unsupervised} implicitly learn to correlate the spatial-spectral information from unregistered multimodality images through an unsupervised framework, and applies to the geometric misaligned images and reference images collected from a different time and sources,
Zheng \etal \cite{zheng2021nonregsrnet} propose a NonRegSRNet that considers more complex misalignment by randomly shifting some pixels of the aligned reference. Nevertheless, most of these methods are still limited at deal with complex misalignments and they are often restricted to unsupervised approaches due to the lack of real-world unaligned datasets. As a result, the fusion-based HSI super-resolution with real unaligned reference images is still under-explored for real-world dual hybrid camera systems. 

In this paper, we explore the fusion-based HSI super-resolution (also dubbed as HSI Fusion) with real-world RGB reference images that have both rigid and non-rigid misalignments. As shown in Figure \ref{fig:demo}, the RGB reference images under our system share the same scene as LR HSI but are not necessarily to be well-aligned. Therefore, we can easily build a dual-camera system using a common commercial tripod to capture paired data, without any special equipment (\eg, beam splitter) as \cite{fu2019hyperspectral}. This makes our approach more economically and technically practical for real-world applications.
In order to effectively address the complex misalignment in real HSI-RGB pairs, we propose an HSI fusion network (HSIFN) with heterogenous feature extractions, multi-stage feature alignments, and attentive feature fusion.
Specifically, the input HSI and RGB images are first transformed into two sets of multi-scale features with an HSI encoder and an RGB encoder, respectively. Then, the features of RGB reference images are processed by a multi-stage alignment module to explicitly align the features of RGB reference with the LR HSI. 
Different from previous works \cite{fu2020simultaneous, nie2020unsupervised} that assumes a global rigid geometric transformation, our alignment module performs the pixel-wise transformation by estimating a dense optical flow map for each level of reference features, which makes our model more robust to non-rigid deformation.
Moreover, the aligned features of RGB reference are adaptively adjusted with an element-wise weight map, which is computed by fusing the features of RGB reference, LR HSI, and predicted optical flow.
This allows our network to selectively focus on more informative regions from the RGB reference while ignoring the incorrect ones, \eg, falsely aligned regions by the previous alignment module. Finally, we combine the aligned features from RGB reference images and LR HSI through a multi-level HSI decoder to generate the reconstructed HR HSI.

Although the misalignment has been a long-standing issue for HSI fusion \cite{zhou2019integrated,fu2020simultaneous}, it is seldom addressed due to the lack of real unaligned datasets. In order to enable the training and evaluation of the proposed method, we collect a real-world HSI fusion dataset, consisting of unaligned high-resolution HSIs captured by a dual-camera system. Each pair of HSIs share the same scene under different viewpoints, and one of them can be selected to synthesize the multispectral image (MSI) or RGB counterpart for HSI fusion with MSI or RGB guidance. To evaluate the effectiveness of the proposed HSI fusion network, extensive experiments are conducted on both simulated and real-world unaligned datasets. The experimental results show that our method obtains a clear improvement over existing single-image and fusion-based super-resolution methods on quantitative assessment as well as visual comparison.

Our main contributions are summarized as follows. 
\begin{itemize}
	\item We propose an HSI fusion network for the fusion-based HSI super-resolution using real unaligned RGB reference with both rigid and non-rigid transformation. 
	\item We introduce a multi-stage pixel-wise alignment module and an adaptive attention module to address the misalignment between RGB reference and LR HSI.
	\item  We collect an HSI fusion dataset with real unaligned RGB reference for verification, and the experiments demonstrate the proposed method achieve better performance than previous works on the real and simulated datasets. 
\end{itemize}

\section{Related Works}

The methods for HSI super-resolution could generally be divided into single image super-resolution methods and reference-based super-resolution methods. In this section, we provide an overview of their recent major approaches.

\subsection{Single Image Super-Resolution}

Single image super-resolution (SISR) \cite{hu2020hyperspectral,jiang2020learning,fu2021bidirectional,li2020mixed, he2018hyperspectral, li2017hyperspectral, mei2017hyperspectral} has been actively studied in recent years for lifting spatial resolution of HSI. 
Due to the ill-posedness of super-resolution, most existing SISR approaches \cite{hu2020hyperspectral,jiang2020learning} rely on the learning capability of deep convolutional neural network (CNN) to recover the missing high-frequency details. 
For example, Hu \etal \cite{hu2020hyperspectral} present an intrafusion network (IFN) that utilizes the spatial-spectral information in one integrated network. Jiang \etal \cite{jiang2020learning} propose SSPSR that exploits the spatial and spectral prior with group convolution and channel attention. Since both spatial and spectral information is important for HSI, 3D convolution is densely explored. Mei \etal \cite{mei2017hyperspectral} propose a novel three-dimensional full convolutional neural network (3D-FCNN) to exploit both the spatial context of neighboring pixels and the spectral correlation of neighboring bands. Li \etal \cite{li2020mixed} design a mixed convolutional network (MCNet) by mixing 2D and 3D convolutions. Fu \etal \cite{fu2021bidirectional} introduce a bidirectional 3D quasi-recurrent neural network (Bi3DQRNN) to explore the structural spatial-spectral correlation and global correlation along spectra. 
To explicitly enforce the constraints on the spatial and spectral domain, He \etal \cite{he2018hyperspectral} propose a deep Laplacian pyramid network to progressively increase the spatial resolution of HSI, whose spectral characteristics are further enhanced by non-negative dictionary learning. Li \etal \cite{li2017hyperspectral} present a deep spectral difference convolutional neural network (SDCNN) model to learn the mapping between LR HSI and HR HS and a spatial constraint (SCT) strategy. With the development of vision transformer \cite{dosovitskiy2020image}, recent works \cite{liu2022interactformer,guo2022external} also explore the transformer architecture to better enhance the modeling abilities for long-range dependency.
Despite the promising performance these methods have achieved, it is physically difficult for SISR to recover highly textured regions because of the information bottleneck of LR input.

\begin{figure*}[t]
  \centering
  \includegraphics[width=0.99\linewidth]{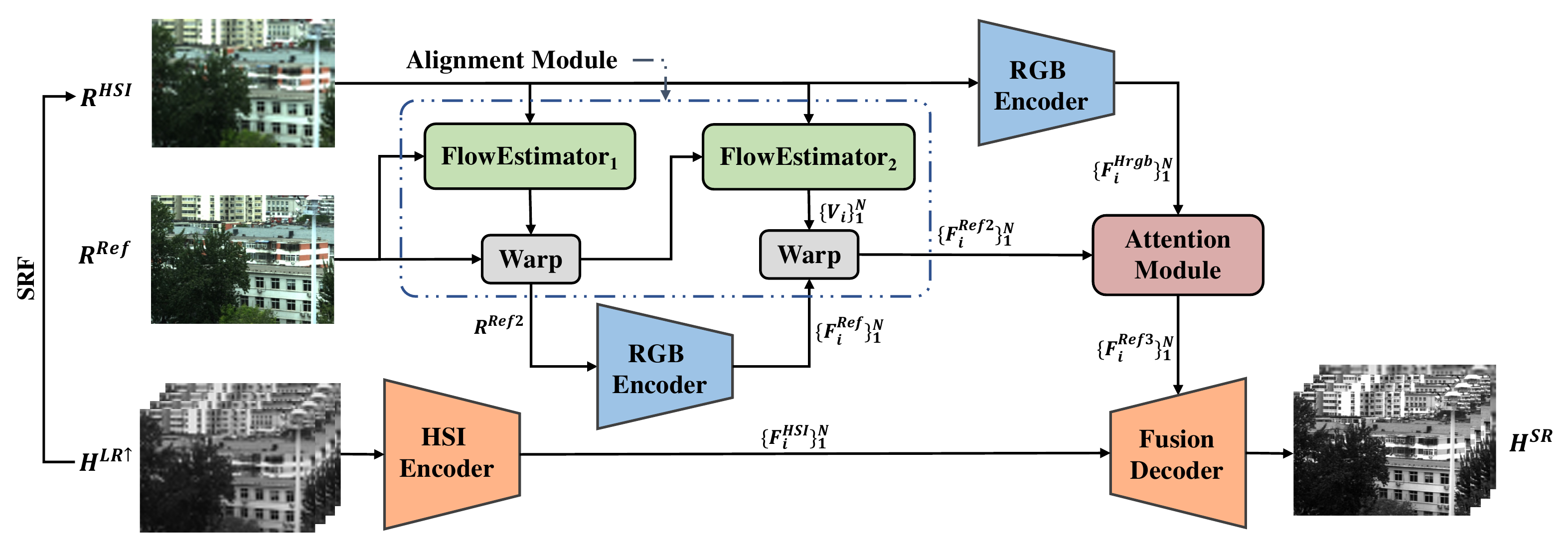}

  \caption{An overview of our unaligned HSI fusion network. It takes LR HSI $H^{LR\uparrow}$, HR reference RGB image $R^{Ref}$, and synthetic RGB image of LR HSI $R^{HSI}$ as inputs. We first extract the multi-level features of the input HSI and RGB images with an HSI encoder and an RGB encoder. Then, we align the reference features $\{F^{Ref}_i\}^N_1$ using optical flow estimated from two successive coarse-to-fine flow estimators. After that, the aligned reference features $\{F^{Ref2}_i\}^N_1$ at different levels are further adjusted by an attention module. Finally, we fuse the weighted aligned feature with the features of LR HSI $\{F^{HSI}_i\}^N_1$ using a fusion decoder to produce the final SR HSI $H^{SR}$.}

  \label{fig:arch}
\end{figure*}

\subsection{HSI Fusion}

Apart from the SISR approaches, another type of fusion-based HSI super-resolution methods \cite{akhtar2014sparse, dong2016hyperspectral, kawakami2011high, kwon2015rgb, akhtar2015bayesian} propose to utilize paired high-resolution RGB or multispectral reference to reconstruct the missing high-frequency details. Previous works following this paradigm usually assume that the paired reference is precisely aligned, which is the key difference between these works and ours. To effectively incorporate the information from the reference, these methods adopt either optimization techniques (\eg matrix factorization \cite{akhtar2014sparse, dong2016hyperspectral, kawakami2011high, kwon2015rgb}, Bayesian representation \cite{akhtar2015bayesian,akhtar2016hierarchical}, and tensor factorization \cite{li2018fusing,zhang2018exploiting}), or deep CNN \cite{dian2018deep,qu2018unsupervised,fu2019hyperspectral}. 
For instance, Akhtar \etal \cite{akhtar2015bayesian} propose to learn the spectral dictionary by using a non-parametric Bayesian model, and apply the dictionary for HSI super-resolution with a generic Bayesian sparse coding strategy. 
Dong \etal \cite{dong2016hyperspectral} formulate the HSI super-resolution as a joint estimation of the hyperspectral dictionary and the sparse codes based on the prior knowledge of the spatial-spectral sparsity of the hyperspectral image. Dian \etal \cite{dian2018deep} proposes to refine the optimization-based fusion framework with a learned deep convolutional neural network-based prior. 
Fu \etal \cite{fu2019hyperspectral} present a simple and efficient CNN to replace the hand-crafted prior for HSI fusion in an unsupervised way. 
Xie \etal \cite{xie2019multispectral} unfold the iteration-based fusion algorithm and propose to learn the proximal operators and model parameters through a deep CNN. 
Despite the superiority of these methods over SISR, their performance is limited for real-world applications due to the requirement of accurate alignment. 

To alleviate the strong assumption of precise alignment, recent works \cite{fu2020simultaneous, nie2020unsupervised, zhou2019integrated} attempt to design models to handle simulated unaligned data. Specifically, Fu \etal \cite{fu2020simultaneous} and Nie \etal \cite{nie2020unsupervised} propose to register images by estimating the geometric transformation matrix via the alternative minimization algorithm and a spatial transformer network, respectively. However, these methods only work on data with simple geometric misalignment, and cannot handle the non-rigid transformation that is more common in real unaligned data. Zhou \etal \cite{zhou2019integrated} propose an integrated registration and fusion method for remote sensing datasets, but it suffers from significant performance drop for natural hyperspectral data as it is reported in \cite{qu2021unsupervised}. 
Qu \etal \cite{qu2021unsupervised} propose an unsupervised framework that implicitly learns to correlate the spatial-spectral information from unregistered multimodality images, and applies to the geometric misaligned images and reference images collected from a different time and sources,
Zheng \etal \cite{zheng2021nonregsrnet} propose a NonRegSRNet that considers more complex misalignment by randomly shifting some pixels of the aligned reference. 
In this work, we consider the HSI fusion with real unaligned RGB guidance. Our architecture is designed to handle complex misalignment of real data and collect the first real HSI fusion dataset for training and evaluation.

\begin{figure*}[h]
\centering
\subfigure[RGB Encoder]{
\includegraphics[width=0.5\textwidth]{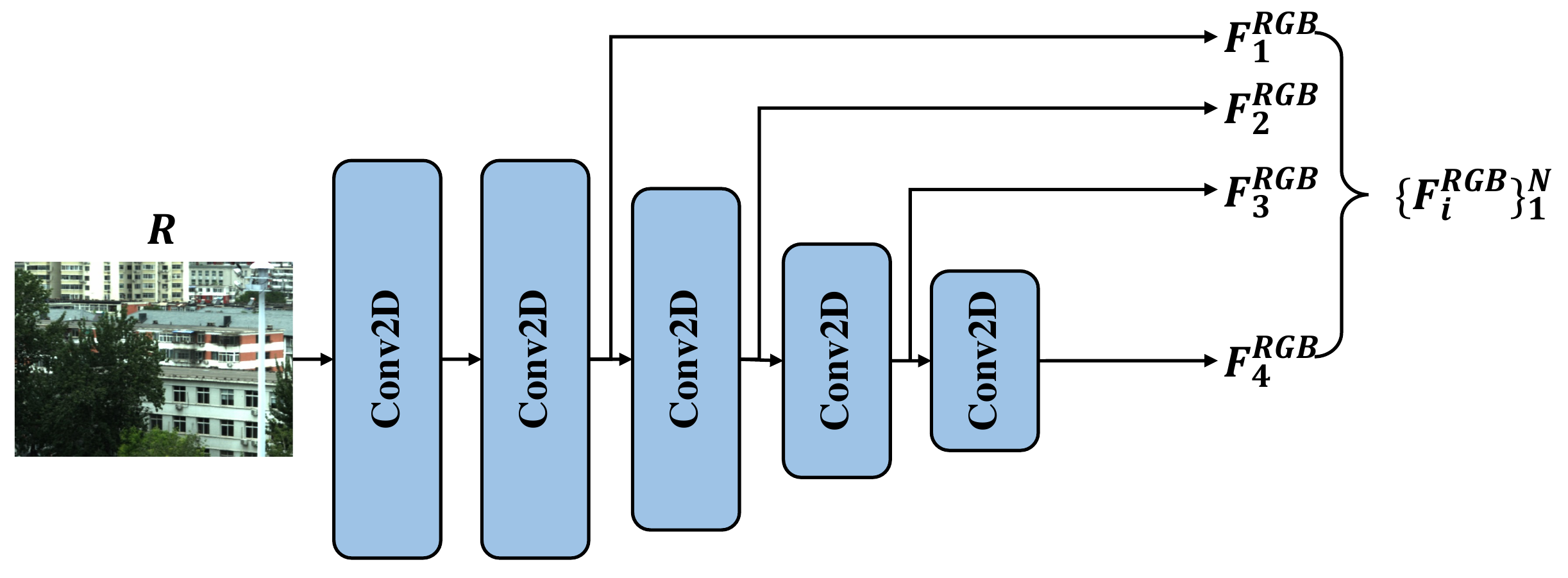}
\label{fig:encoder-rgb}
}
\subfigure[HSI Encoder]{
\includegraphics[width=0.45\textwidth]{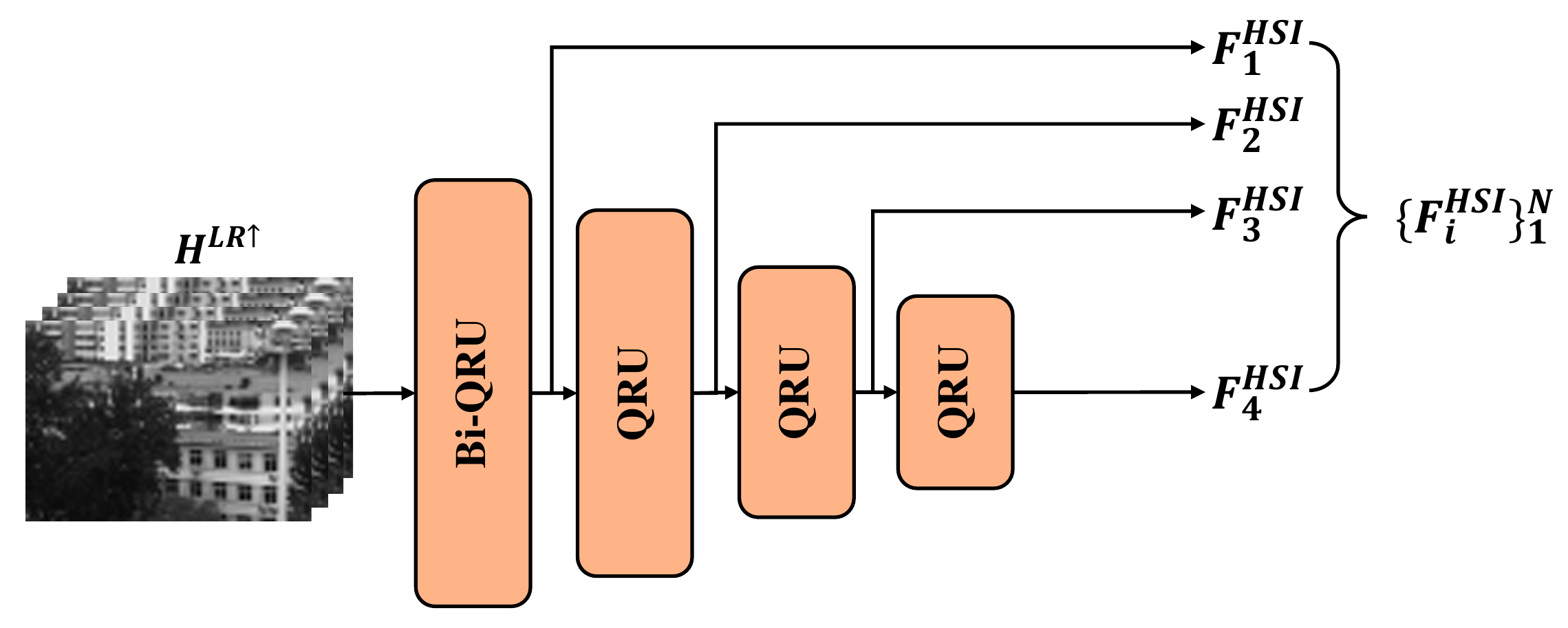}
\label{fig:encoder-hsi}
}
\vspace{-3mm}
\caption{Detailed structures of the RGB encoder and HSI encoder. The RGB encoder contains five layers of convolution, and the HSI encoder is made up of four layers of QRU.}
\end{figure*}

\section{HSI Fusion Network}

Given the input HSI with low spatial resolution $H^{LR}$ and high-resolution reference RGB image $R^{Ref}$, the task of HSI fusion is to reconstruct the high-resolution HSI $H^{SR}$ conditioned on $R^{Ref}$. For the real-world HSI fusion, the precise alignment between reference $R^{Ref}$ and input $H^{LR}$ is usually unattainable due to the high cost and complexity of the required imaging system. Hence, it is essential to properly align the reference with input before performing the fusion.   

To address the aforementioned issue, we propose an HSI fusion network (HSIFN), which consists of three steps, \ie, feature extraction, alignment, and fusion. An overview of our HSIFN is shown in Figure \ref{fig:arch}. It is built with five major components, including an HSI encoder, an RGB encoder, an alignment module, an attention module, and a fusion decoder. Specifically, the two different encoders are responsible for extracting multi-level deep features from input HSI and RGB reference by considering the specific characteristics of each type of image. For each level feature of the reference RGB image, the alignment module estimates a dense optical flow map in a coarse-to-fine manner to perform the pixel-wise warping. After acquiring the aligned reference features, the attention module is employed to compute an element-wise attention weight map to drive the network to attend to more discriminative regions. Then, the weighted aligned features, as well as the HSI features, are further integrated with the fusion decoder to produce the super-resolved (SR) HSI. The details of each network component of our architecture are described in the subsequent sections.

\subsection{HSI \& RGB Encoders}
The encoders embed the reference RGB image $R^{Ref}$ as well as the upsampled HSI $H^{LR\uparrow}$ into multi-level deep features to extract useful information for the subsequent reconstruction. To better explore the specific characteristics from each type of image, \eg, structural spatio-spectral correlation of HSI, we adopt different encoders to extract features of the reference RGB image and HSI as,
\begin{equation}
  \begin{aligned}
    \{F^{HSI}_i\}^N_1 &= \mathbf{E}_{hsi} (H^{LR\uparrow}), \\
    \{F^{Ref}_i\}^N_1 &= \mathbf{E}_{rgb} (R^{Ref}),
  \end{aligned}
  \label{eq:encoder}
\end{equation}
where $N$ is the total number of levels of features and is set to 4 in our network, $\mathbf{E}_{hsi}$ and $\mathbf{E}_{rgb}$ are HSI and RGB encoders, $F^{HSI}_i$ and $F^{Ref}_i$ is the extracted features of HSI and RGB reference at $i^{th}$ level.

\subsubsection{RGB Encoder} 
We construct the RGB encoder by stacking a series of 2D convolutional layers with kernel size $K=5$ to incorporate information within large receptive fields. Following the common practice in \cite{sun2018pwc, tan2020crossnet++}, we reduce the spatial resolution and increase the number of feature channels as the network goes deeper to extract multi-level features. The detailed network structure is shown in Figure \ref{fig:encoder-rgb}. The RGB encoder consists of five convolution-activation layers. The first two layers keep the spatial dimension, while the last three half both the height and width sequentially. We use the same feature numbers activation function, and scaled exponential linear unit (SELU) \cite{klambauer2017self} activation function, across all the layers.
We use the same RGB encoder to extract features of both reference RGB image $R^{Ref}$ as well as the synthetic RGB image $R^{HSI}$ of input LR HSI $H^{LR\uparrow}$ as
\begin{equation}
	\begin{aligned}
		\{F^{Ref}_i\}^N_1 &= \mathbf{E}_{rgb} (R^{Ref}), \\
		\{F^{Hrgb}_i\}^N_1 &= \mathbf{E}_{rgb}(R^{HSI}),
	\end{aligned}
\end{equation}
where $N$ denotes the number of levels of the multi-level features and is set to 4 in our network.

\subsubsection{HSI Encoder}

The HSI encoder is built in a similar way as the RGB counterpart except that it uses quasi recurrent convolutional unit (QRU) \cite{wei20203}. Different from conventional convolutional layers, QRU uses a 3D convolutional neural network and a recurrent pooling function to better explore the structural spatio-spectral correlation and global correlation along the spectrum for HSI. 
In detail, QRU first separately performs two 3D convolutions on the input features $I$ to obtain a set of pixel-wise weight maps $W$ and candidate feature maps $F$ for each band as 
\begin{equation}
\begin{aligned}
W &= \sigma \left(h_w \otimes I \right), \\
F &= \tanh \left(h_f \otimes I \right),
\end{aligned}
\label{eq:grc-conv}
\end{equation}
where $\sigma$ denotes sigmoid function, $h_w, h_f$ are two 3D filters, and $\otimes$ represents the 3D convolution. Then, the candidate feature map of each band $f_i$ is adaptively merged using the weight map $w_i$ in a recurrent manner as
\begin{equation}
h_i = (1-w_i) \odot h_{i-1} + w_i \odot f_i, \,\,\, \forall  i\in [1,b],
\label{eq:grc-rnn}
\end{equation}
where $\odot$ denotes the element-wise multiplication and $h_i$ denotes the fused feature map at the $i^{th}$ band. The final features $H$ are the concatenation of $h_i$ for each band.

Following \cite{wei20203}, we use the bidirectional QRU (Bi-QRU) for the first layer and the alternating direction scheme is adopted for the subsequent QRU layers. Specifically, the bidirectional QRU essentially computes two sets of features in the forward and backward direction and takes the summation of them as the final features. The alternating direction scheme makes two successive QRU layers merge the features in different directions. These enhance our HSI encoder with the global spectral context without too much computational burden. 

\subsection{Alignment Module}

The key to HSI fusion is to design an effective approach to transfer the high-frequency information from HR RGB reference into LR HSI. When the images are properly aligned, this can be achieved with a simple concatenation of reference and HSI features. However, the same method might be unsuitable for unaligned reference due to the difficulty for subsequent convolutional fusion layers to properly capture the correspondence between features of the RGB reference and LR HSI at different spatial locations.   

With the aim of reducing the negative effect of spatial misalignment for the subsequent reconstruction, we introduce a pixel-wise alignment module to explicitly align the multi-level reference features of RGB reference to the features of LR HSI. Unlike previous works \cite{fu2020simultaneous, nie2020unsupervised} that align the reference with a global rigid homography transformation, our alignment module performs pixel-wise transformation by estimating a dense optical flow map for each level of reference features, which makes our model more robust to non-rigid deformation. 

The overall structure of the proposed alignment module is shown in Figure \ref{fig:arch}. Without sacrificing the representation capability, we first convert the input HSI $H^{LR\uparrow}$ to a synthetic RGB counterpart $R^{HSI}$ (dubbed as HSI-RGB) with a spectral response function (SRF) for computational efficiency. To better handle large displacements, following the previous works \cite{ilg2017flownet, tan2020crossnet++}, we perform the optical flow estimation in a coarse-to-fine manner using two successive flow estimators. Specifically, the first flow estimator $ \mathbf{Flow_1}$ takes reference RGB $R^{Ref}$and HSI-RGB $R^{HSI}$ as input and predicts a rough flow map to coarsely align the reference,
\begin{equation}
	R^{Ref2} = \mathbf{warp}(R^{Ref}, \mathbf{Flow_1}(R^{Ref}, R^{HSI})).
\end{equation}
Then, the coarsely-aligned reference $R^{Ref2}$ and HSI-RGB $R^{HSI}$ are fed into the second flow estimator $\mathbf{Flow_2}$ to predict a set of refined flow maps to align multi-level reference features, 
\begin{equation}
\begin{aligned}
\{V_i\}_1^N &= \mathbf{Flow_2}(R^{Ref2}, R^{HSI}), \\
\{F^{Ref2}_i\}^N_1 &= \mathbf{warp}(\{F^{Ref}_i\}^N_1, \{V_i^N\}).
\end{aligned}
\end{equation}

Although any state-of-the-art optical flow networks \cite{hui2018liteflownet,sun2018pwc,dosovitskiy2015flownet,ilg2017flownet} can be adopted as our flow estimator, not everyone can be effectively trained under our architecture without explicit supervision from ground truth optical flow. Hence, we empirically choose different networks for datasets with/without sufficient training samples to achieve the best performance. Specifically, an improved version of FlowNetS \cite{zheng2018crossnet} is adopted and trained from scratch for large datasets (\eg, our simulated dataset), and a pretrained PWC-Net \cite{sun2018pwc} is used for small datasets (\eg, our real dataset).

\begin{figure}[t]
  \centering
  \includegraphics[width=1\linewidth]{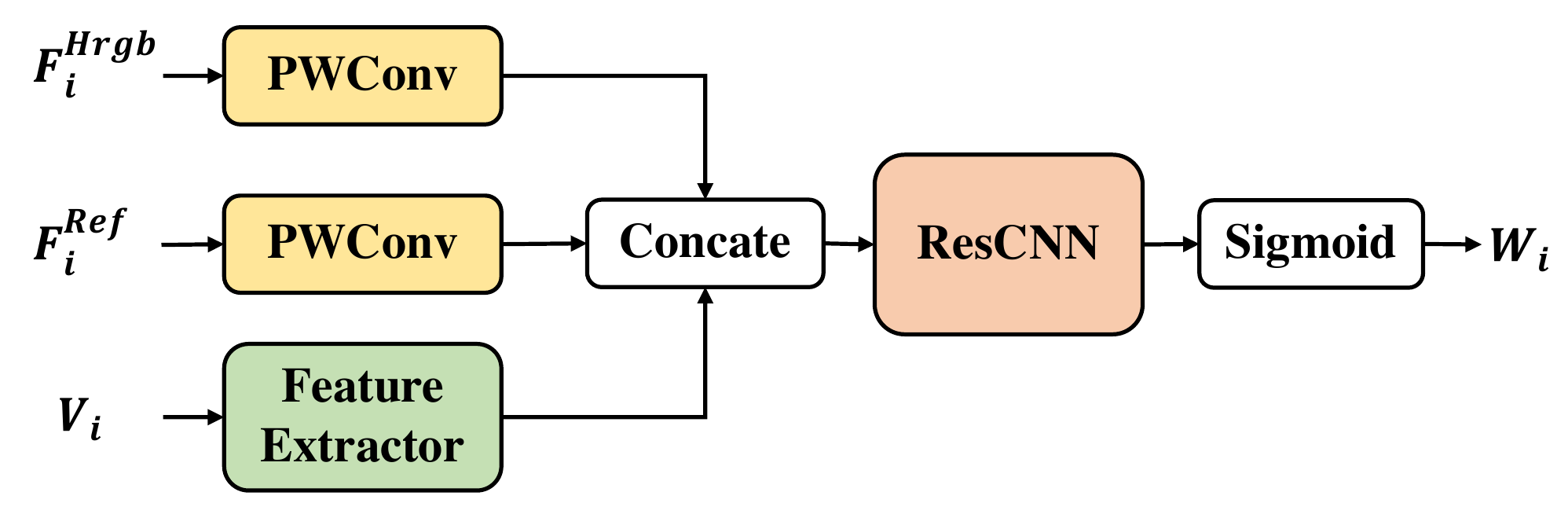}

  \caption{Illustration of the computation of attention map. The optical flow $V_i$ is embedded into deep features and the point-wise convolution (PWConv) is used to lower the dimensions of the features of HSI-RGB $F_i^{Hrgb}$ and reference RGB $F_i^{Ref}$.}
  \label{fig:attention}
\end{figure}

\subsection{Attention Module}

Despite the previous alignment module being able to align the reference features to some extent, mistakes of flow estimation are unavoidable and can be even more common for HSI fusion due to the lack of ground truth optical flow for explicit guidance. Furthermore, the warping operation itself also introduces misleading ghosting artifacts in the occluded area \cite{zhao2020maskflownet}, which are useless and cause confusion in the subsequent fusion step.   

On the basis of analysis, we introduce an attention module to adaptively adjust the importance of each spatial location in the feature map by computing an element-wise attention weight. Before performing attention, we first encode the HSI-RGB $R^{HSI}$ into multi-level deep features $\{F^{Hrgb}_i\}^N_1$, using the same RGB encoder as RGB reference,
\begin{equation}
	\{F^{Hrgb}_i\}^N_1 = \mathbf{E}_{rgb}(R^{HSI}).
\end{equation}
Then, the estimated optical flow, the features of HSI-RGB, and the RGB reference are fed into an attention module to predict the corresponding attention map. 

The structure of the proposed attention module is shown in Figure \ref{fig:attention}. The predicted optical flow $V_i$ is first fed into a feature extractor $\mathbf{E}_{f}$, \ie, a small CNN, to obtain the embedded flow features $F^{flow}_i$, 
\begin{equation}
	F^{flow}_i = \mathbf{E}_{f} (V_i).
\end{equation}
Meanwhile, a point-wise convolutional layer $\mathbf{G}$ is used to lower the dimensions of HSI and RGB features for computational efficiency. Then, we concatenate the flow features, the compressed HSI and RGB features along the channel dimension and feed the results into a residual convolutional network (ResCNN) to obtain the initial attention weight, which is subsequently normalized with the sigmoid function $\sigma$,
\begin{equation}
	W_i = \sigma(\mathbf{ResCNN}([\mathbf{G}(F^{Ref}_i), \mathbf{G}(F^{Hrgb}_i), F^{flow}_i])).
\end{equation}

Afterward, the normalized weight is element-wise multiplied with the aligned reference features $F^{Ref2}_{i}$ to produce the final reference features for the fusion decoder, 
\begin{equation}
	F^{Ref3}_{i} = W_i \otimes F^{Ref2}_{i}.
\end{equation}

\begin{figure}[t]
  \centering
  \includegraphics[width=1\linewidth]{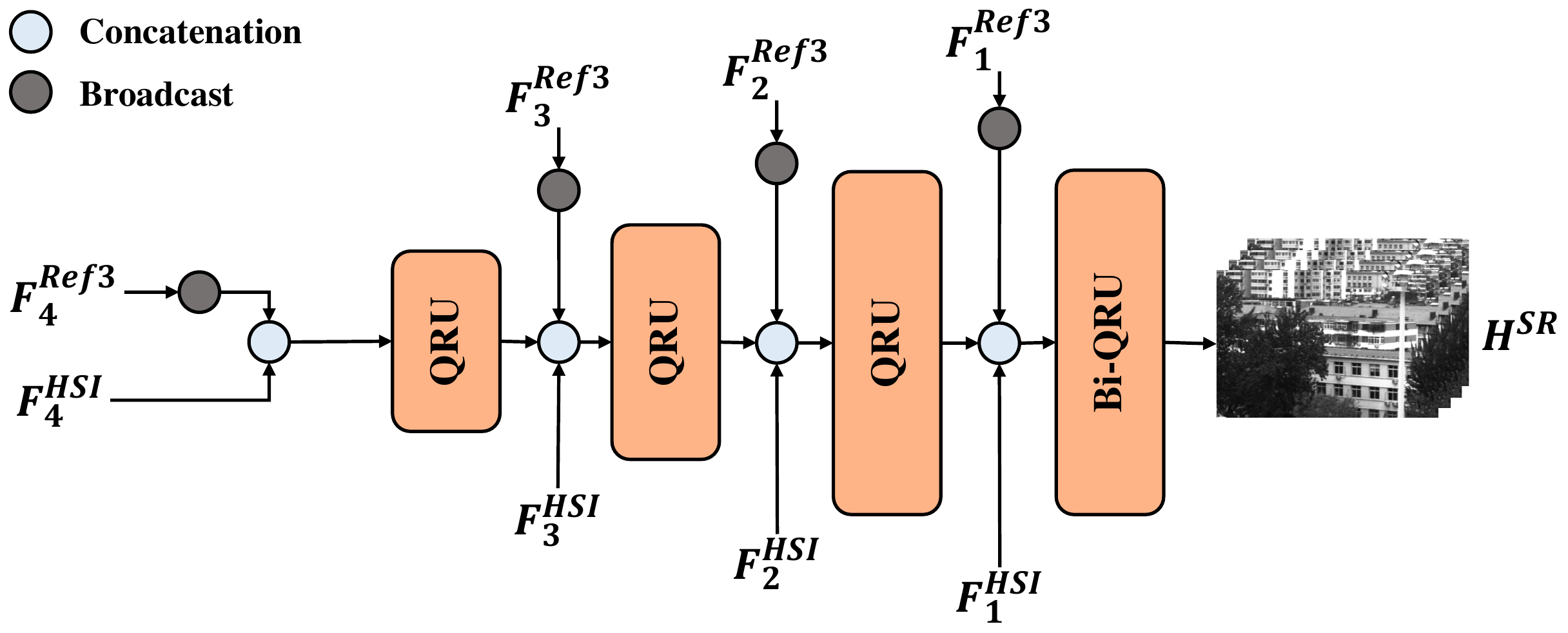}
  \vspace{-5mm}
  \caption{Illustration of the fusion decoder. $F_i^{Ref3}$ denotes the weighted aligned reference features at level $i$. $F_i^{HSI}$ denotes the HSI features at level $i$.}
  \label{fig:decoder}
\end{figure}

\begin{figure*}[t]
  \centering
  \includegraphics[width=1\linewidth]{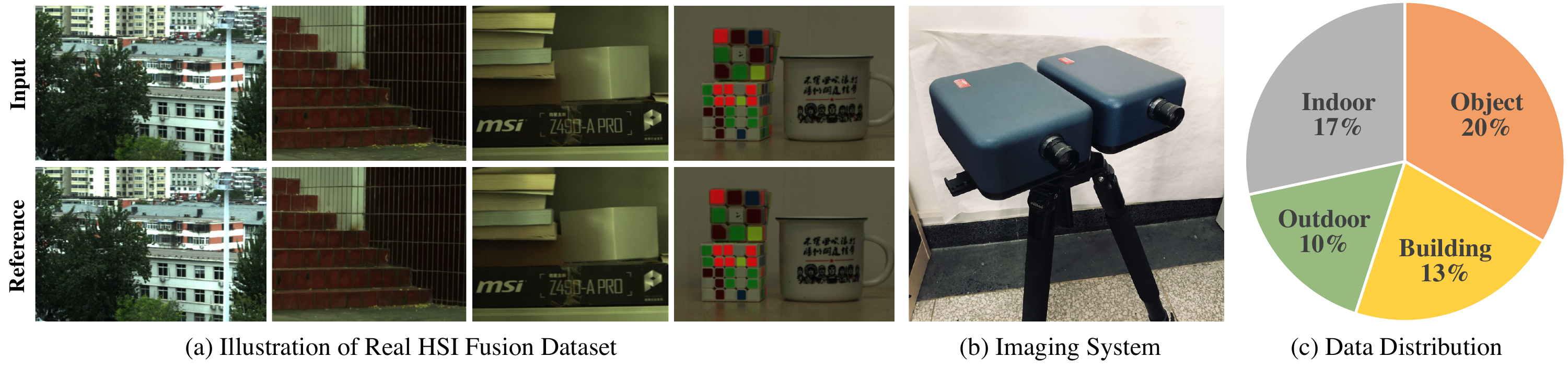}

  \caption{(a) \textbf{Illustration of Real HSI Fusion Dataset.} The dataset includes paired unaligned HSIs from indoor scenes, outdoor scenes, objects, and buildings. The first line and the second line show the synthetic RGB images from captured HSIs. \textbf{(b) Imaging System.} The imaging system is made up of two HSI cameras mounted on a tripod. \textbf{(c) Data Distribution.}}

  \label{fig:dataset}
\end{figure*}

\subsection{Fusion Decoder}

With the weighted aligned reference features, a fusion decoder with the skip connection \cite{ronneberger2015u} is adopted to predict the final high-resolution HSI. In detail, the multi-level features of RGB reference are the first broadcast along the band dimension to match the number of band of the input LR HSI. Then, the expanded reference features, HSI features at level $i$, and the decoder features at level $i-1$ (if exist) are concatenated and sent to an upsample-QRU \cite{wei20203} layer to produce the decoder features at level $i$ (for $i\in[0,2]$). After obtaining the decoder features at the last level, another bidirectional-QRU  \cite{wei20203} layer is employed to perform the last fusion to generate the final SR output.

The illustration of the fusion decoder is shown in Figure \ref{fig:decoder}. Specifically, the fusion decoder contains four upsampled QRU layers \cite{wei20203} to integrate the features of reference RGB image and LR HSI at four levels. The upsampled QRU is identical to the QRU described in the previous section of the HSI encoder except it uses upsampled 3D convolution \cite{odena2016deconvolution} instead of plain 3D convolution. Similar to the HSI encoder, the last QRU layer is bidirectional and the rest are alternative directional. For QRU layer at level $i$, it receives the concatenation of decoder features $F_{i-1}^{D}$ at level $i-1$ (if exists), the broadcasted reference features $F^{Ref3}_{4-i}$ and HSI features $F_{4-i}^{HSI}$ at level $4-i$, and predicts the decoder features at next level $i+1$, \ie, 
\begin{equation}
	F_i^{D} = QRU([F_{i-1}^{D}, F_{4-i}^{HSI}, broadcast(F^{Ref3}_{4-i})]).
\end{equation} 
For the last QRU layer, it predicts the final SR HSI.

\section{Real HSI Fusion Dataset} \label{sec:dataset}

Most existing works on HSI fusion \cite{li2018fusing,fu2019hyperspectral,fu2020simultaneous, nie2020unsupervised} either assume the precise alignment between the reference image and input HSI or only consider the simulated unaligned data generated by geometric transformation. The performance and the generalization capabilities of these methods on real data are usually not taken into account due to the lack of appropriate real unaligned HSI fusion datasets. 

In order to validate the performance of our approach, we collect a new real-world unaligned HSI fusion dataset, called Real-HSI-Fsuion, which consists of 60 pairs of unaligned high-resolution HSIs. An overview of the dataset is shown in Figure \ref{fig:dataset}, our dataset contains different types of scenes including indoor scenes, outdoor scenes, buildings, and objects. Each pair of HSIs shares the same scene under different viewpoints, but are not precisely aligned. 

In detail, we employ two SOC710-VP hyperspectral cameras from Surface Optics Corporation (SOC), USA, for the HSI imaging. Each camera is equipped with a silicon-based charge-coupled device (CCD) and an integrated scanning system to capture HSI with 696 $\times$ 520 pixels in spatial resolution and 128 spectral bands from 376.76 $nm$ to 1037.77 $nm$ at 5.16 $nm$ interval. The dynamic range of each HSI is 12 bits, so the spectral value ranges from 0 to 4095. We use a commercial dual camera mount tripod to fix the two cameras as shown in Figure \ref{fig:dataset}. Due to the lack of autofocus, we manually adjust the exposure time, focal length, and camera position to maximize the clarity and overlapped region for each scene. 

After acquiring the raw HSI data, we coarsely align the image pairs by estimating the affine transformation matrix with SIFT \cite{lowe1999object} and RANSAC \cite{fischler1981random} using the synthetic RGB counterparts. Then, we manually crop the overlapped region for each pair to remove the disjoint border. Following the common practice in \cite{arad2016sparse, chakrabarti2011statistics}, we select 31 bands ranging from about 400 nm to 700 nm in visible spectral range to construct the final dataset that consists of 60 HR HSI-HSI pairs, which share a similar size as existing HSI datasets, \eg, CAVE and Harvard. Ten pairs of HSIs are randomly selected for testing and the rest is for training.  

It should be noted that the dataset consists of pairs of HSIs captured by two HSI cameras. The complementary RGB image is synthesized from one HSI by multiplying it with the spectral response matrix of Nikon D700 as \cite{qu2021unsupervised, akhtar2015bayesian,lanaras2015hyperspectral}. It might be confusing why not directly capture RGB reference with RGB camera directly. The reasons behind such a choice are mainly to make the dataset more flexible and useful for other slightly different settings without collecting similar datasets again. For example, by using our dataset, it is possible to generate different synthetic RGB references using different camera response functions. Besides, it is also feasible to synthesize an unaligned multispectral reference image (MSI) for HSI SR with an unaligned MSI reference. Further, our dataset can also be used for spectral super-resolution with unaligned LR HSI reference where two HSIs are required.

\section{Experiments}

In this section, we provide the experimental results on both the simulated dataset and our real HSI fusion dataset. We also provide an ablation study and discussion to verify the effectiveness of each proposed network component.

\begin{table*}[h]
\centering
\setlength{\tabcolsep}{0.3cm}
\caption{Quantitative comparison on simulated unaligned dataset for $4\times$ and $8\times$  HSI SR with reference. Different from previous work, our simulated dataset is generated from real unaligned RGB-RGB pairs. }
\label{tab:simulate} 
\begin{tabular}{@{}cccccccccccc@{}}
\toprule
 \makecell{Scale\\ Factor}    &  \makecell{Metric\\ ~}    & \makecell{Bicubic \\ ~} & \makecell{Bi-3DQRNN \\ \cite{fu2021bidirectional}}  & \makecell{SSPSR \\ \cite{jiang2020learning}}  & \makecell{MCNet \\ \cite{li2020mixed}}   & \makecell{NSSR \\ \cite{dong2016hyperspectral}}     & \makecell{Optimized \\ \cite{fu2019hyperspectral}}     & \makecell{Integrated \\ \cite{zhou2019integrated}} & \makecell{NonReg \\ \cite{zheng2021nonregsrnet}}  & \makecell{u2MDN \\ \cite{qu2021unsupervised}}    & \makecell{HSIFN \\ (Ours)}  \\ \midrule
\multirow{3}{*}{$\times 4$} & PSNR $\uparrow$   & 29.88   & 37.53     & 37.02 & 37.35 & 27.26   & 25.40   & 29.09 & 25.92 & 25.85  & \textbf{42.06} \\
                            & SSIM $\uparrow$   & 0.914   & 0.979     & 0.976 & 0.979 & 0.875   & 0.848   & 0.908 & 0.838 & 0.840  & \textbf{0.991}       \\
                            & SAM $\downarrow$  & 0.055   & 0.030     & 0.045 & 0.034 & 0.065   & 0.319   & 0.232 & 0.311 & 0.127  & \textbf{0.026}       \\ \midrule
\multirow{3}{*}{$\times 8$} & PSNR $\uparrow$   & 25.49   & 30.12     & 30.22 & 30.03 & 25.92   & 25.42   & 25.97 & 25.56 & 25.39  & \textbf{38.29} \\
                            & SSIM $\uparrow$   & 0.835   & 0.921     & 0.925 & 0.921 & 0.852   & 0.963   & 0.869 & 0.823 & 0.834 & \textbf{0.983}       \\
                            & SAM $\downarrow$  & 0.094   & 0.057     & 0.057 & 0.057 & 0.077   & 0.318   & 0.312 & 0.323 & 0.150 & \textbf{0.036}       \\ 
\bottomrule
\end{tabular}

\end{table*}

\begin{figure*}[]
 \centering
 \setlength{\tabcolsep}{0.09cm}
  
 \begin{tabular}{cccccc}

  \includegraphics[height=0.075\textwidth]{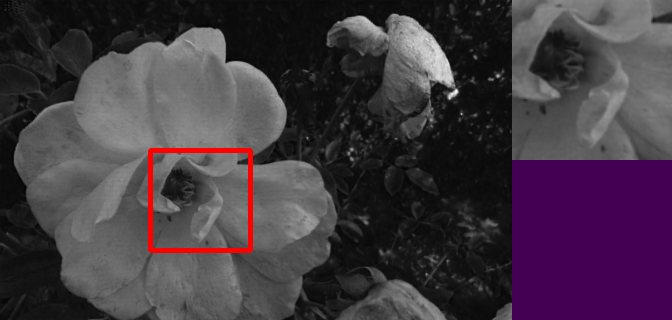}
  & \includegraphics[height=0.075\textwidth]{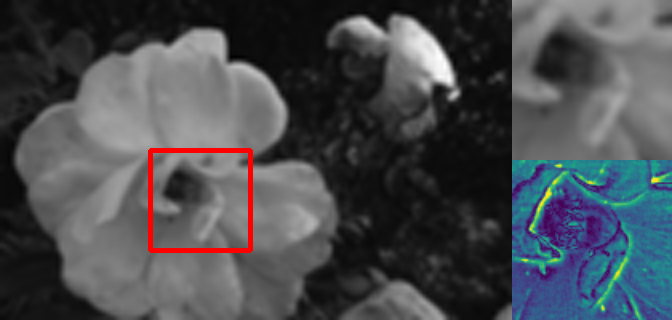}
  & \includegraphics[height=0.075\textwidth]{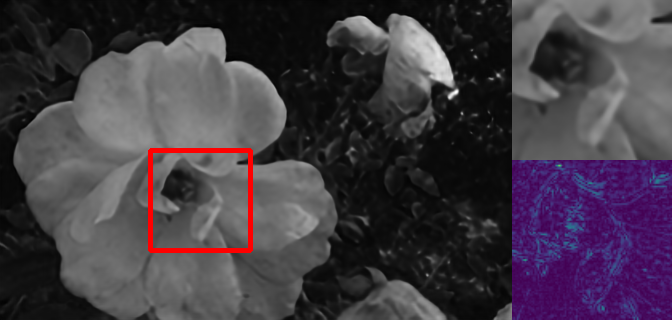}
  & \includegraphics[height=0.075\textwidth]{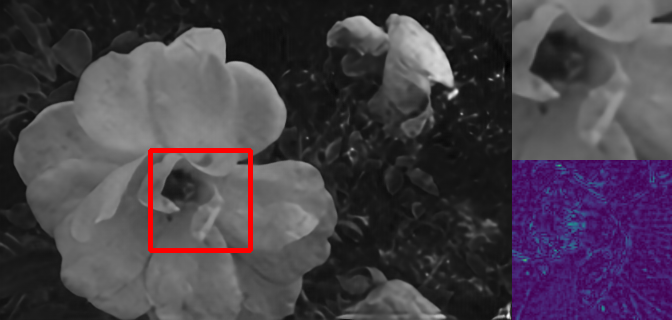}
  & \includegraphics[height=0.075\textwidth]{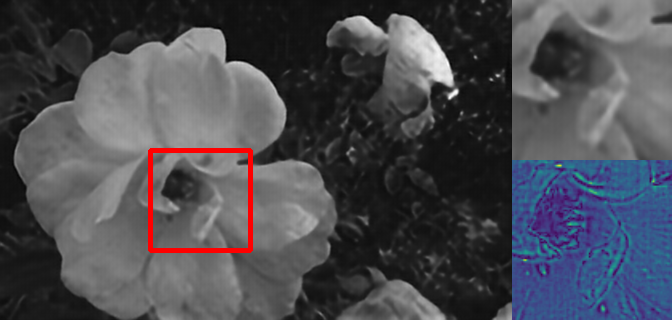}
  & \includegraphics[height=0.075\textwidth]{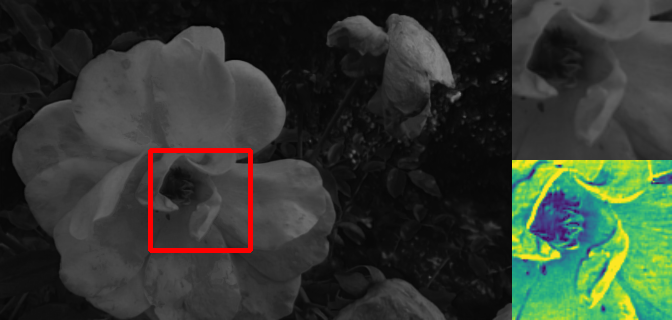}\\
  Ground Truth  & Bicubic & Bi3DQRNN \cite{fu2021bidirectional} & SSPSR \cite{jiang2020learning} & MCNet \cite{li2020mixed} & NSSR \cite{dong2016hyperspectral} \\
  
  \includegraphics[height=0.075\textwidth]{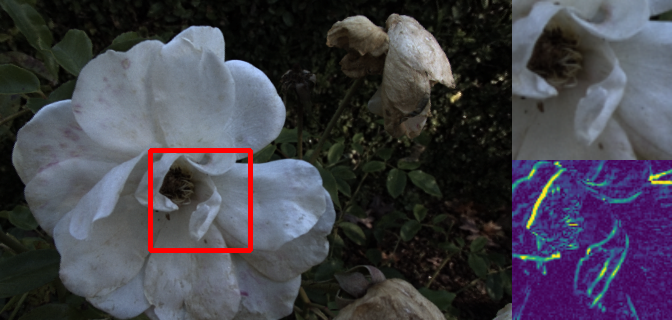}
  & \includegraphics[height=0.075\textwidth]{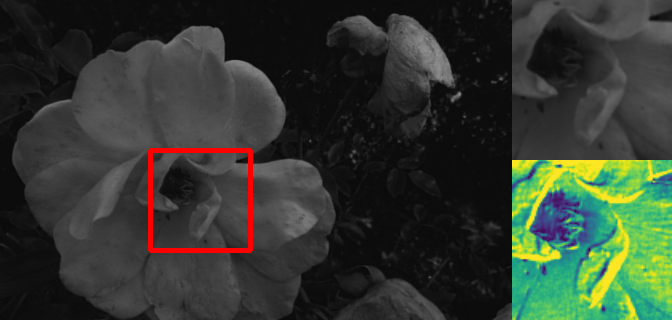}
  & \includegraphics[height=0.075\textwidth]{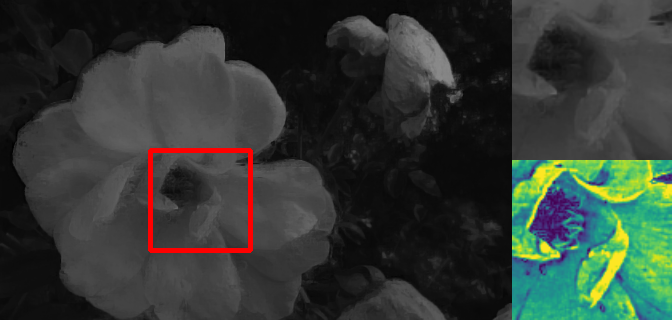}
  & \includegraphics[height=0.075\textwidth]{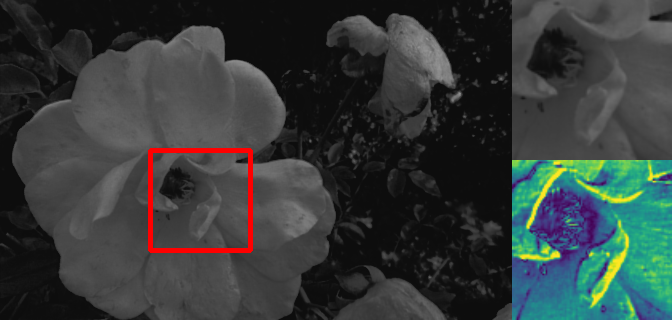}
  & \includegraphics[height=0.075\textwidth]{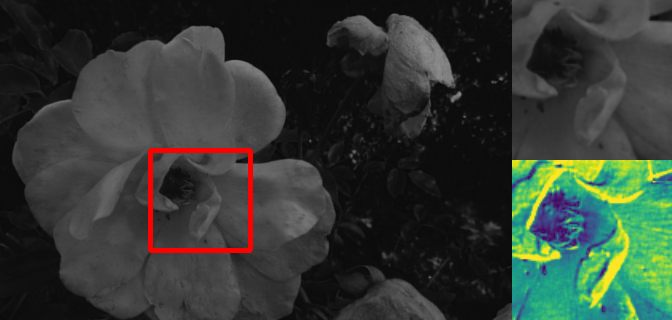}
  & \includegraphics[height=0.075\textwidth]{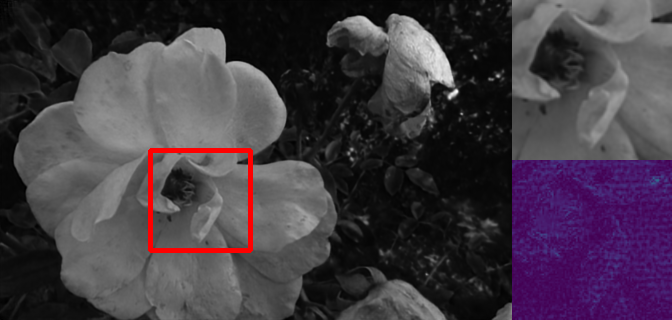}\\
  Reference ($\times4$) & Optimized \cite{fu2019hyperspectral} & Integrated \cite{zhou2019integrated} & NonReg \cite{zheng2021nonregsrnet} & u2MDN \cite{qu2021unsupervised} & HSIFN (Ours) \\
  
  \hline\\ 
  
  \includegraphics[height=0.075\textwidth]{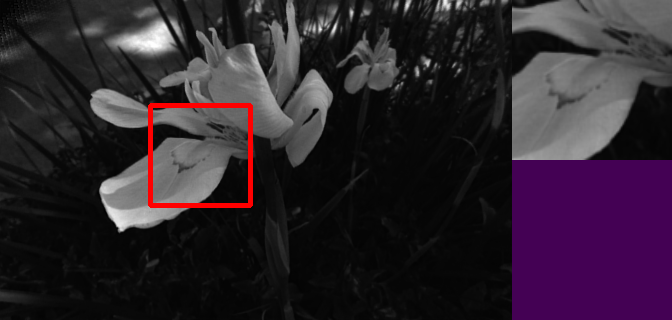}
  & \includegraphics[height=0.075\textwidth]{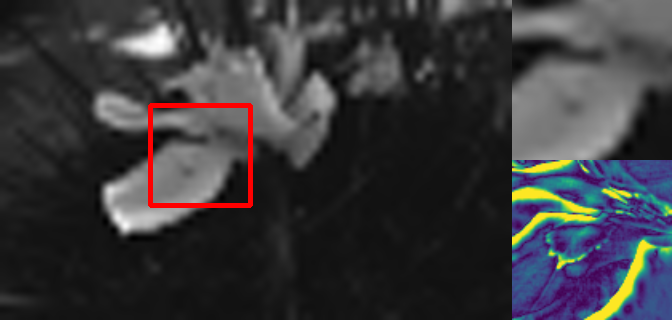}
  & \includegraphics[height=0.075\textwidth]{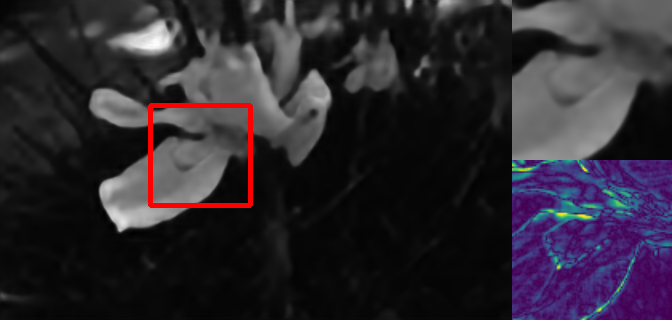}
  & \includegraphics[height=0.075\textwidth]{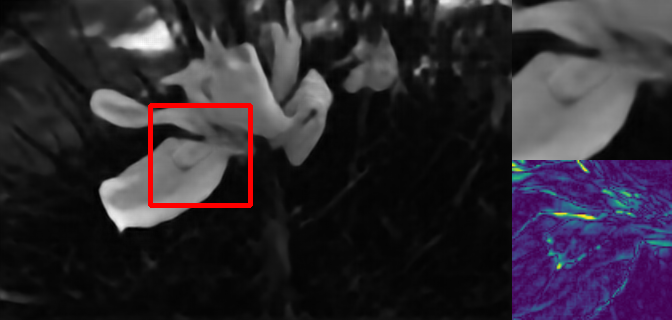}
  & \includegraphics[height=0.075\textwidth]{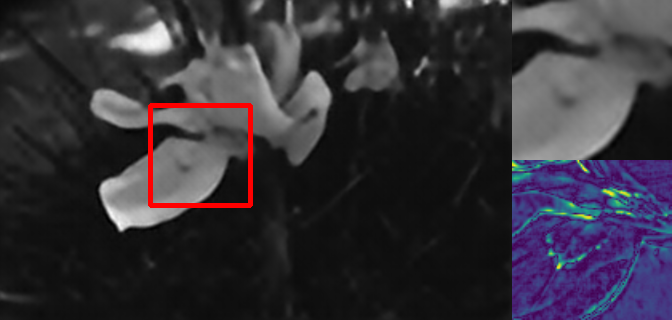}
  & \includegraphics[height=0.075\textwidth]{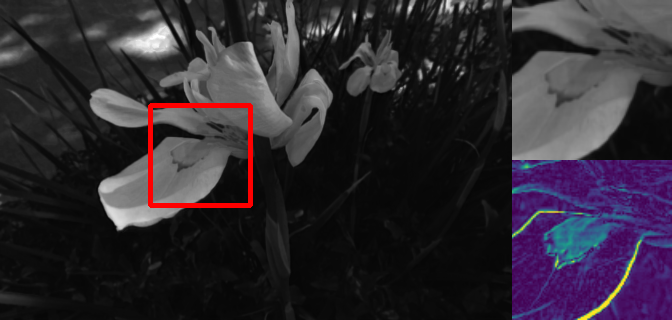}\\
  Ground Truth  & Bicubic & Bi3DQRNN \cite{fu2021bidirectional} & SSPSR \cite{jiang2020learning} & MCNet \cite{li2020mixed} & NSSR \cite{dong2016hyperspectral} \\
  
  \includegraphics[height=0.075\textwidth]{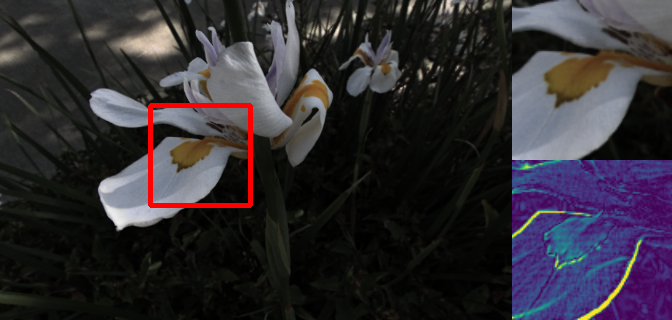}
  & \includegraphics[height=0.075\textwidth]{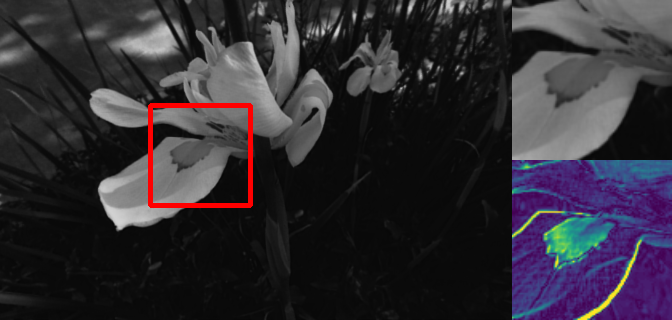}
  & \includegraphics[height=0.075\textwidth]{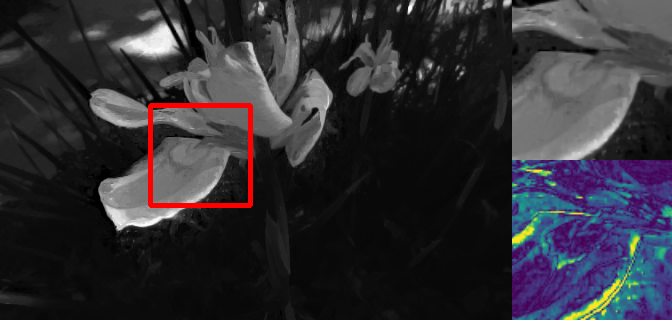}
  & \includegraphics[height=0.075\textwidth]{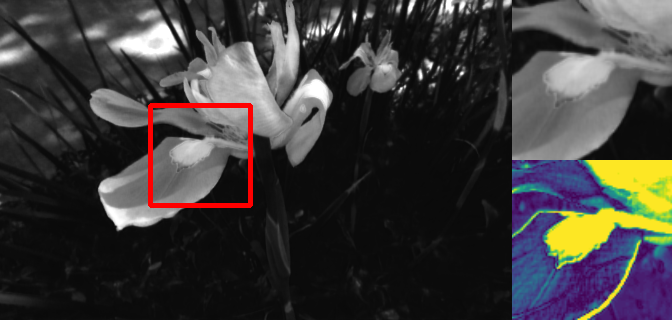}
  & \includegraphics[height=0.075\textwidth]{imgs/results/flower/x4/u2mdn}
  & \includegraphics[height=0.075\textwidth]{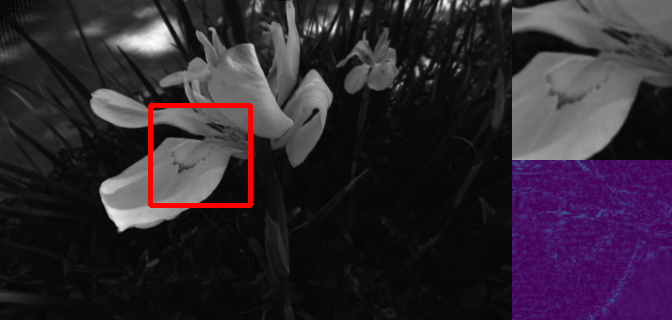}\\
Reference ($\times8$) & Optimized \cite{fu2019hyperspectral} & Integrated \cite{zhou2019integrated} & NonReg \cite{zheng2021nonregsrnet} & u2MDN \cite{qu2021unsupervised} & HSIFN (Ours) \\
	
 \end{tabular}

 \caption{Visual comparison on simulated unaligned dataset under scale factors of 4 and 8. Our method produces sharper details than competing SISR approaches and more aligned results over fusion-based methods. Zoom in for details.}

 \label{fig:flower-vis} 
\end{figure*}

\subsection{Experimental Settings}

\subsubsection{Dataset} To evaluate the proposed method under different levels of misalignment, we perform the experiments on the simulated dataset (with small misalignment) as well as the real dataset we collected (with relatively larger misalignment). Different from previous works \cite{fu2020simultaneous, nie2020unsupervised}, we construct the simulated dataset by synthesizing the HSIs with 31 bands from real unaligned RGB-RGB image pairs in the light field dataset Flowers \cite{srinivasan2017learning}, using HSCNN+ \cite{shi2018hscnn}, which is a recent state-of-the-art deep-learning model for hyperspectral recovery from RGB images. The simulated dataset contains 3343 pairs of images in the size of $320 \times 512$ where 100 pairs are randomly selected for testing and the rest is used for training.
 For the real dataset, 10 pairs of images are randomly selected for testing as  described in Section \ref{sec:dataset}. For each HSI-HSI pair, we choose one HSI from them to synthesis the RGB reference (Ref-RGB), and use the same approach to synthesis the RGB counterpart of input HSI (HSI-RGB). 
 Due to the difference between two HSI cameras, an extra histogram-based color-matching is performed to alleviate the spectral inconsistency. 
 We generate the LR HSI for simulated and real datasets using the same approach as \cite{fu2021bidirectional}, \ie, the HR image is first blurred using a Gaussian kernel with $ \mu=8, \sigma=3$ and then downsampled with the specified scale factor.

\subsubsection{Implementation Details} We implement the proposed fusion network using PyTorch \cite{paszke2019pytorch}. The AdamW \cite{loshchilov2017decoupled} optimizer is adopted to minimize the smooth $L_1$ loss between predicted SR HSI and the corresponding ground truth. The weight decay rate of the optimizer is set to $5\times10^{-5}$. The batch size is set to 1. For the simulated dataset, we train the network for 50 epochs with a learning rate set to $1\times 10^{-4}$. For the real dataset, we train the network for 200 epochs with a learning rate set to $1\times 10^{-5}$. Besides, several strategies are used to improve the performance on the real dataset,  
(1) We use the pretrained PWC-Net \cite{sun2018pwc} as our flow estimator for the real dataset, since training the flow estimator from scratch is extremely difficult without sufficient supervision (inadequate training samples, the lack of ground truth optical flow). 
(2) We rescale the weight of the HSI features and the reference features to balance the importance of each type of feature for different scale factors.
(3) We pretrain our network with the HR HSI-RGB and then fine-tune on the LR HSI-RGB. This allows our network to distill knowledge from the easier HR-HR matching to guide the more ambiguous LR-HR matching. 

\begin{table*}[]
\centering
\setlength{\tabcolsep}{0.3cm}
\caption{Quantitative comparison on our real unaligned HSI fusion dataset for $4\times$ and $8\times$ HSI SR with reference. }
\label{tab:real}
\begin{tabular}{@{}cccccccccccc@{}}
\toprule
 \makecell{Scale\\ Factor}    &   \makecell{Metric\\ ~}   & \makecell{Bicubic \\ ~} & \makecell{Bi-3DQRNN \\ \cite{fu2021bidirectional}}  & \makecell{SSPSR \\ \cite{jiang2020learning}}  & \makecell{MCNet \\ \cite{li2020mixed}}   & \makecell{NSSR \\ \cite{dong2016hyperspectral}}     & \makecell{Optimized \\ \cite{fu2019hyperspectral}}     & \makecell{Integrated \\ \cite{zhou2019integrated}} & \makecell{NonReg \\ \cite{zheng2021nonregsrnet}} & \makecell{u2MDN \\ \cite{qu2021unsupervised}}    & \makecell{HSIFN \\ (Ours)}  \\ \midrule
\multirow{3}{*}{$\times 4$} & PSNR $\uparrow$   & 34.07  & 37.80  & 39.04 & 39.07 & 30.83  & 27.26  & 30.66 & 31.60 & 30.58	& \textbf{41.21} \\
                    & SSIM $\uparrow$   		& 0.941  & 0.969  & 0.976 & 0.974 & 0.958  & 0.916  & 0.935 & 0.951	& 0.936	& \textbf{0.989}       \\
                    & SAM $\downarrow$  		&  0.042 & 0.044  & 0.040 & \textbf{0.038} & 0.050  & 0.189  & 0.157 & 0.081 & 0.087	& 0.047       \\ \midrule
\multirow{3}{*}{$\times 8$} & PSNR $\uparrow$   & 28.43  & 31.66  & 31.13 & 31.31 & 28.02  & 26.99  & 30.97 &24.42  & 30.18	& \textbf{33.13} \\
                    & SSIM $\uparrow$   		& 0.870  & 0.908  & 0.906 & 0.904 & 0.939  & 0.909  & 0.932 & 0.117	& 0.936	& \textbf{0.946}       \\
                    & SAM $\downarrow$  &\textbf{0.058}  & 0.060  & 0.069 & 0.063 & 0.063  & 0.195  & 0.138 & 0.712 & 0.096	& 0.061       \\   
                     \bottomrule
\end{tabular}
\end{table*}

\begin{figure*}[]
 \centering
 \setlength{\tabcolsep}{0.09cm}
  
 \begin{tabular}{cccccc}
  
  \includegraphics[height=0.081\textwidth]{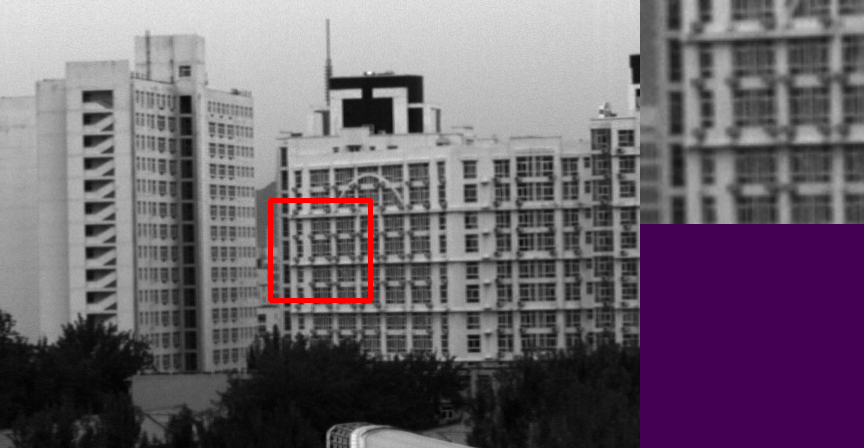}
  & \includegraphics[height=0.081\textwidth]{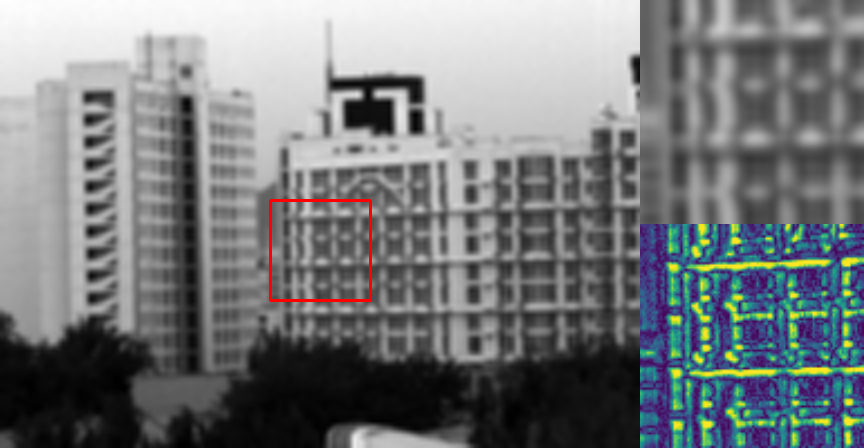}
  & \includegraphics[height=0.081\textwidth]{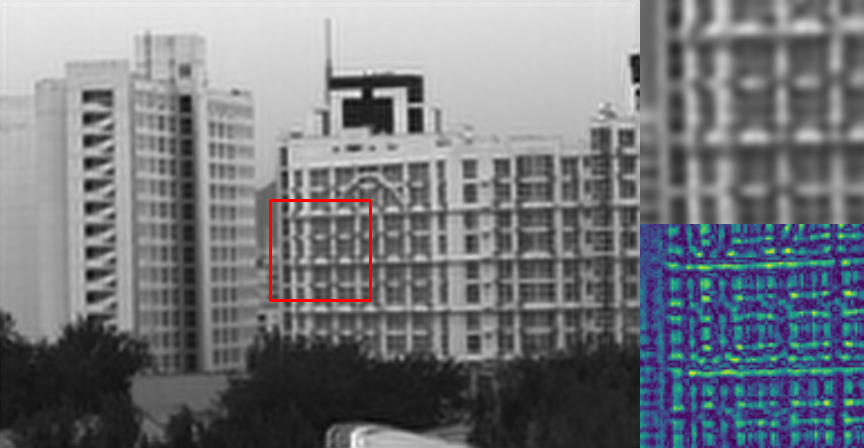}
  & \includegraphics[height=0.081\textwidth]{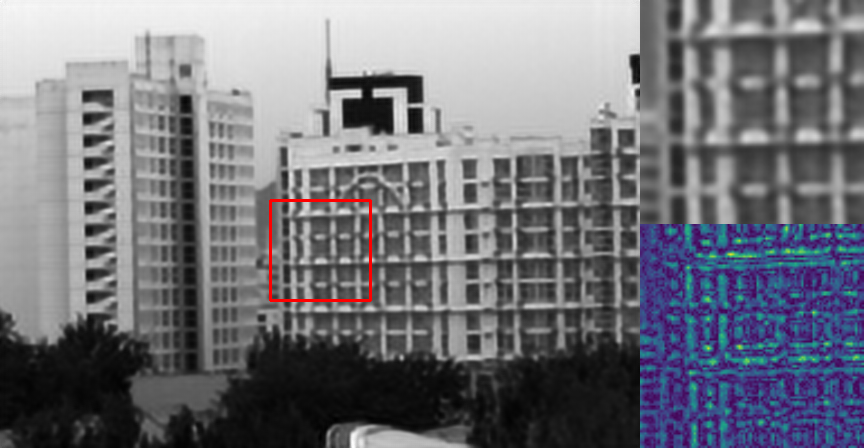}
  & \includegraphics[height=0.081\textwidth]{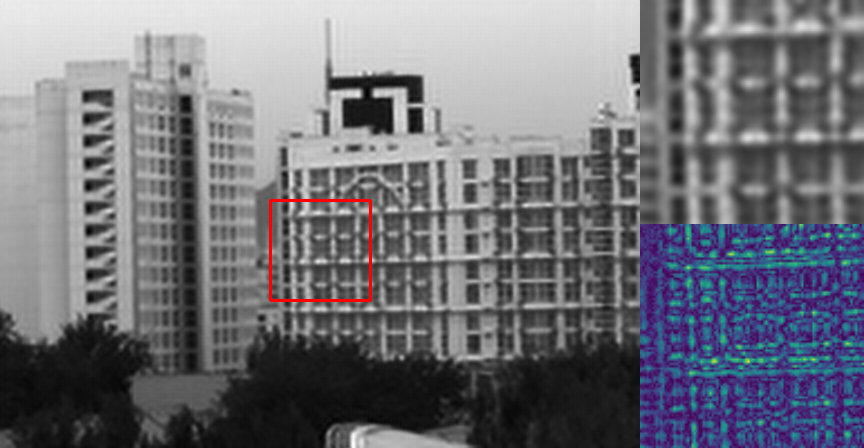}
  & \includegraphics[height=0.081\textwidth]{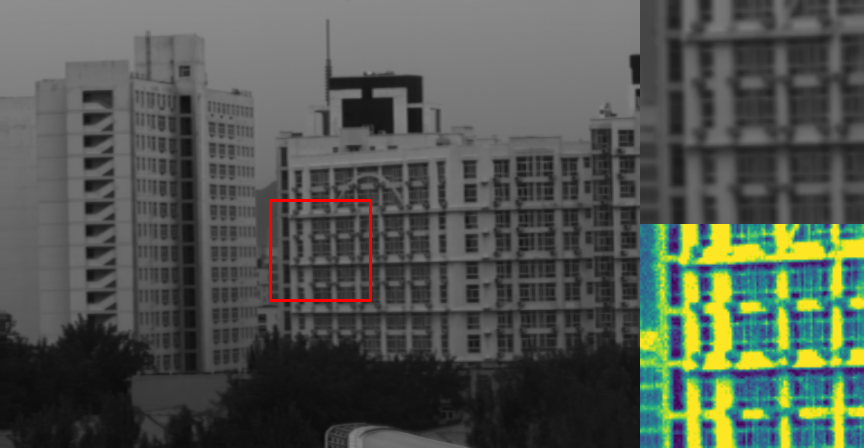}\\
  Ground Truth  & Bicubic & Bi3DQRNN \cite{fu2021bidirectional} & SSPSR \cite{jiang2020learning} & MCNet \cite{li2020mixed} & NSSR \cite{dong2016hyperspectral} \\
  
  \includegraphics[height=0.081\textwidth]{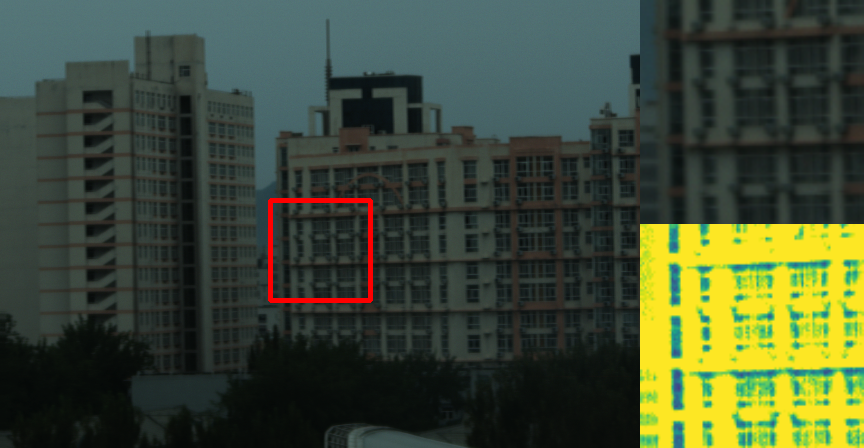}
  & \includegraphics[height=0.081\textwidth]{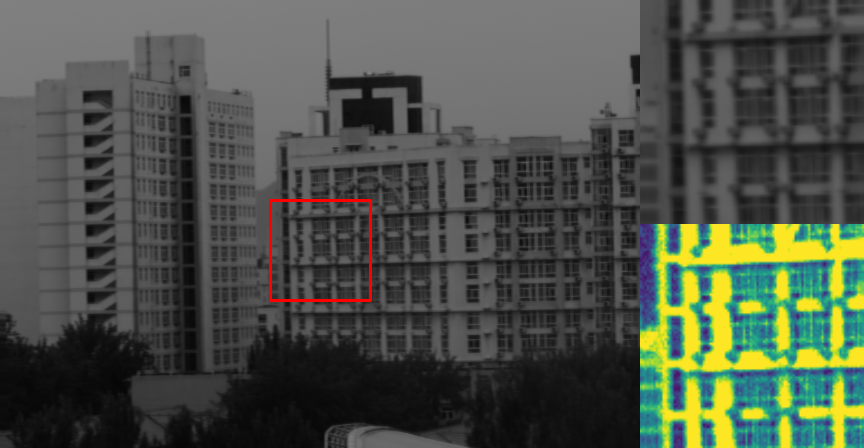}
  & \includegraphics[height=0.081\textwidth]{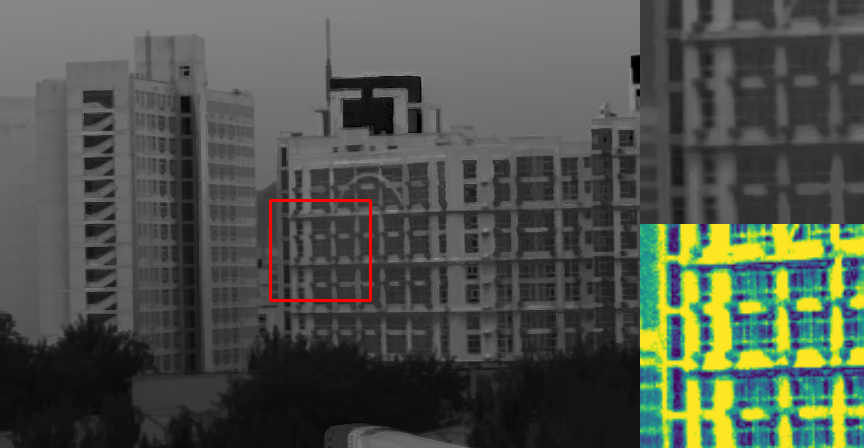}
  & \includegraphics[height=0.081\textwidth]{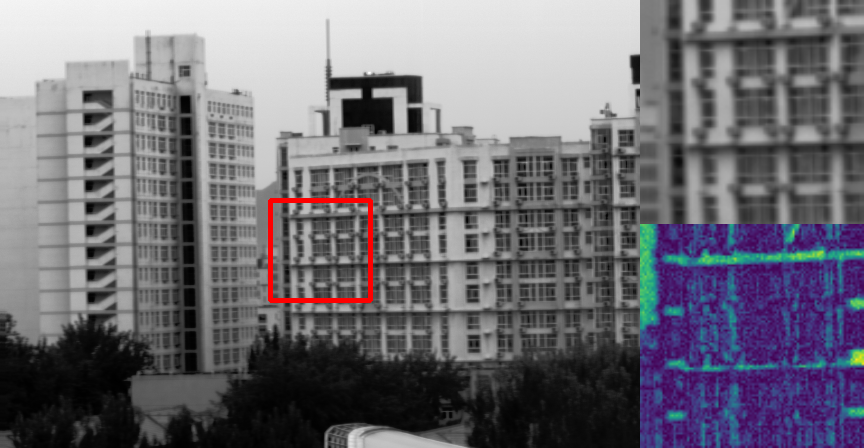}
  & \includegraphics[height=0.081\textwidth]{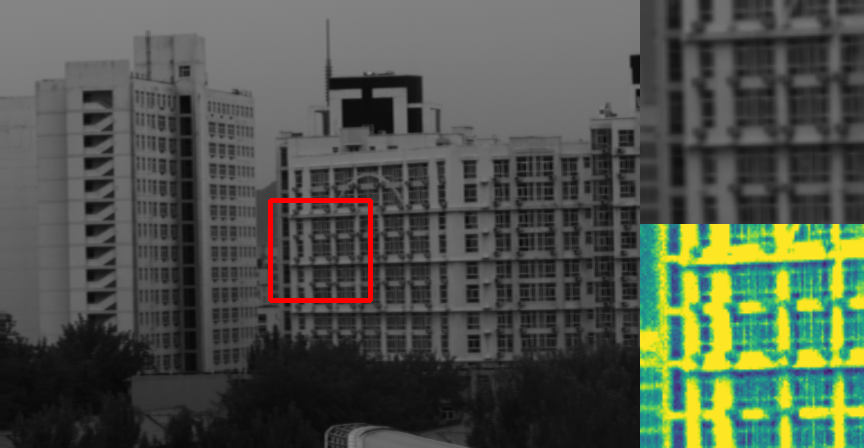}
  & \includegraphics[height=0.081\textwidth]{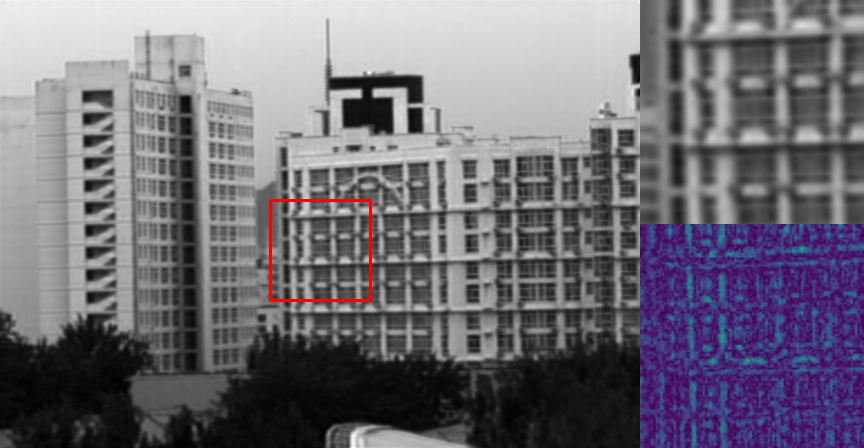}\\
Reference ($\times4$) & Optimized \cite{fu2019hyperspectral} & Integrated \cite{zhou2019integrated} & NonReg \cite{zheng2021nonregsrnet} & u2MDN \cite{qu2021unsupervised} & HSIFN (Ours)  \\
  
  \hline\\ 
  
  \includegraphics[height=0.081\textwidth]{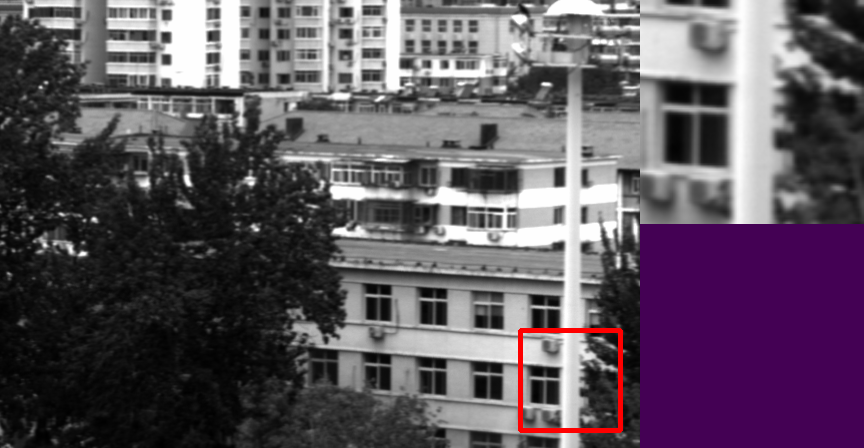}
  & \includegraphics[height=0.081\textwidth]{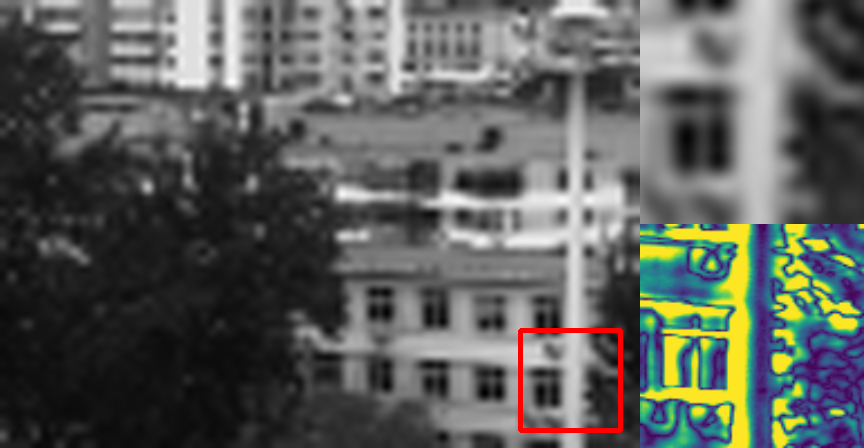}
  & \includegraphics[height=0.081\textwidth]{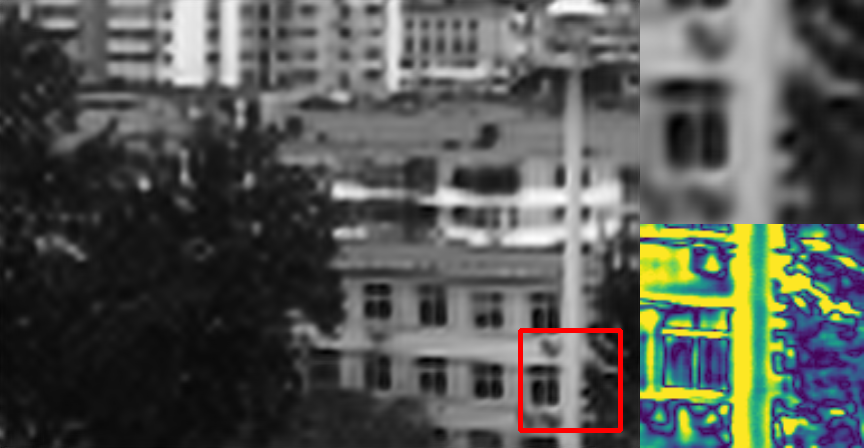}
  & \includegraphics[height=0.081\textwidth]{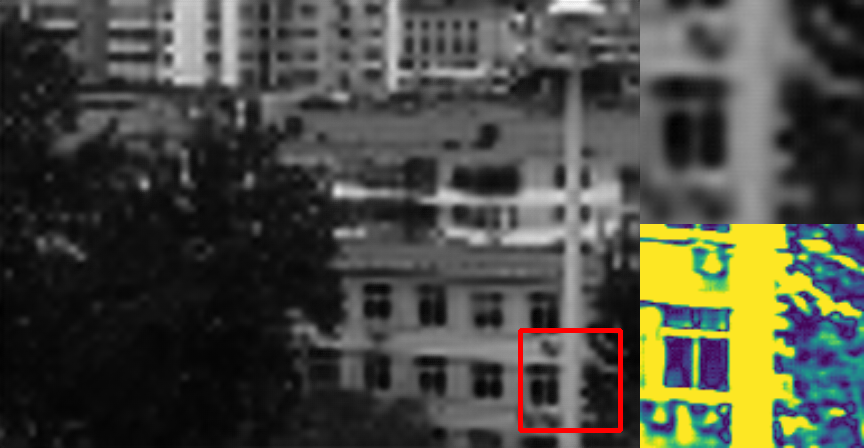}
  & \includegraphics[height=0.081\textwidth]{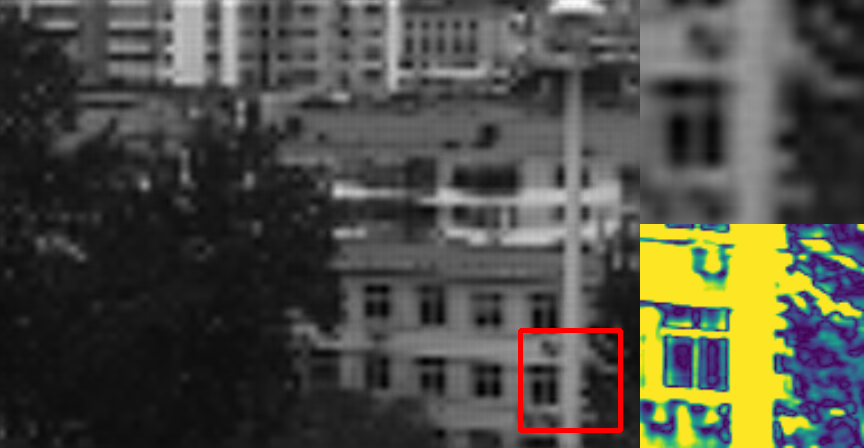}
  & \includegraphics[height=0.081\textwidth]{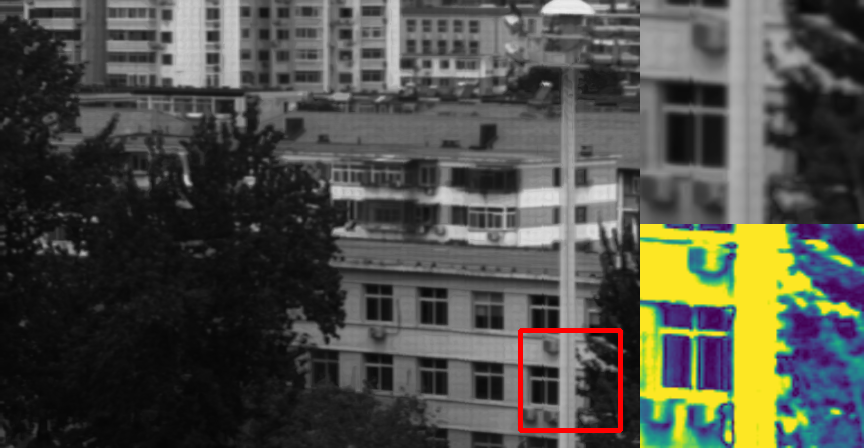}\\
  Ground Truth  & Bicubic & Bi3DQRNN \cite{fu2021bidirectional} & SSPSR \cite{jiang2020learning} & MCNet \cite{li2020mixed} & NSSR \cite{dong2016hyperspectral}  \\
  
  \includegraphics[height=0.081\textwidth]{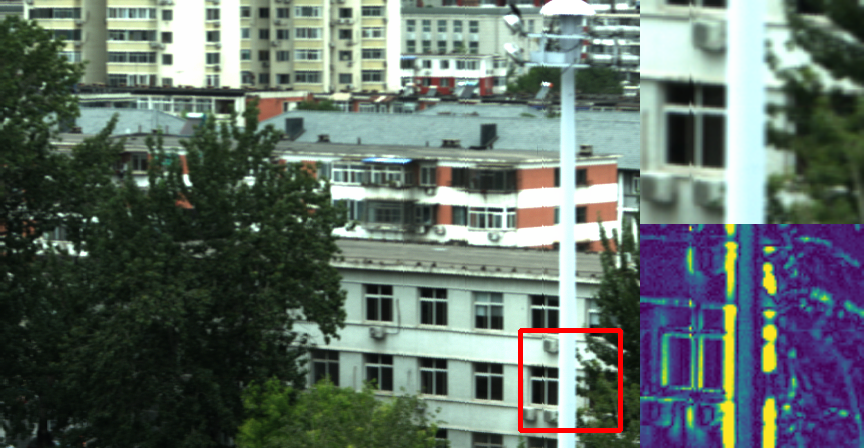}
  & \includegraphics[height=0.081\textwidth]{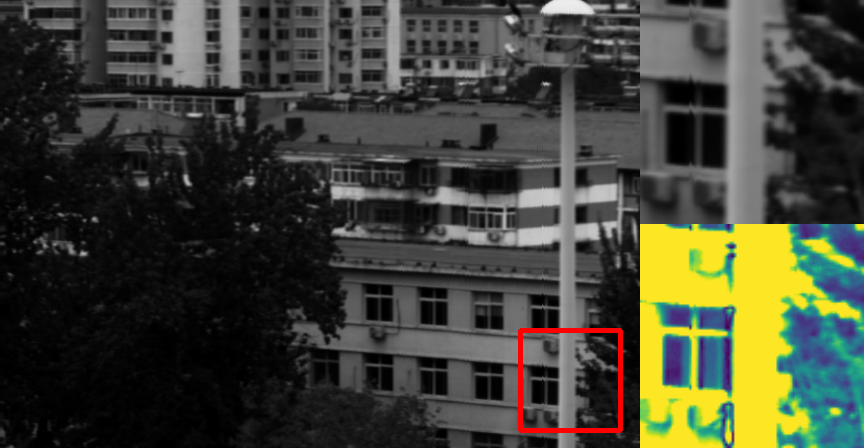}
  & \includegraphics[height=0.081\textwidth]{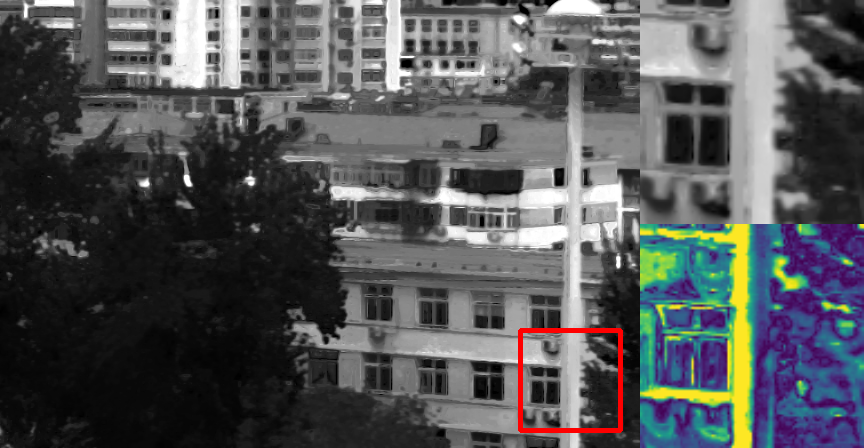}
  & \includegraphics[height=0.081\textwidth]{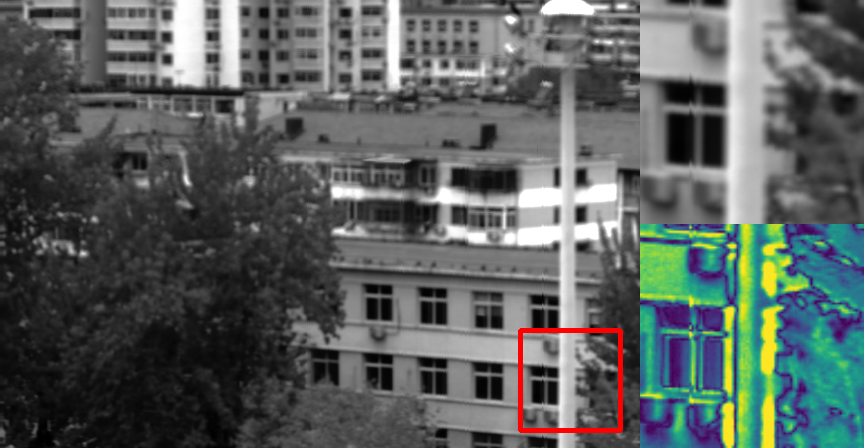}
  & \includegraphics[height=0.081\textwidth]{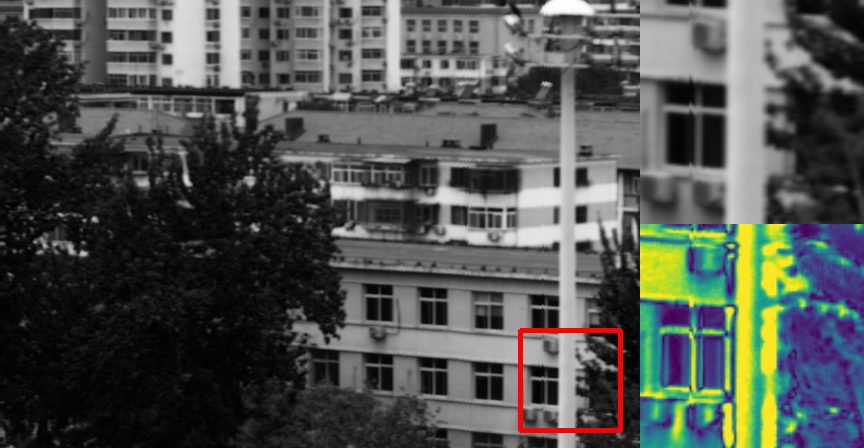}
  & \includegraphics[height=0.081\textwidth]{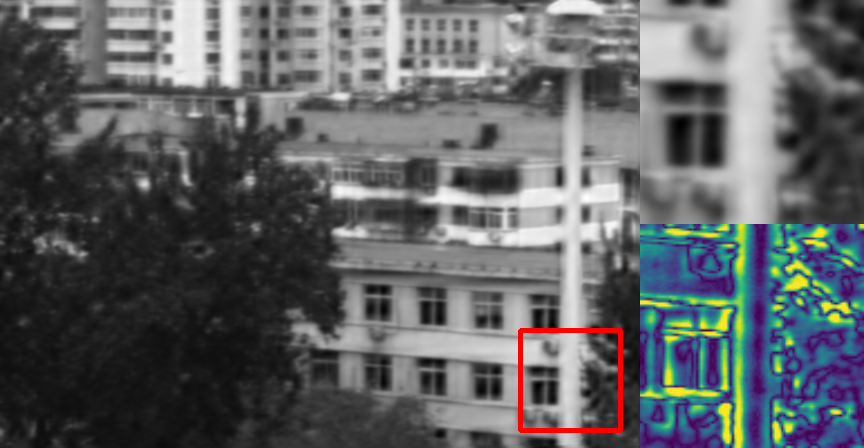}\\
Reference ($\times8$) & Optimized \cite{fu2019hyperspectral} & Integrated \cite{zhou2019integrated} & NonReg \cite{zheng2021nonregsrnet} & u2MDN \cite{qu2021unsupervised} & HSIFN (Ours) \\

 \end{tabular}

 \caption{Visual comparison on our real HSI fusion dataset under scale factors of 4 and 8. Our method reconstructs sharper details while properly aligning with the ground truth. Zoom in for details.}

 \label{fig:real-vis} 
\end{figure*}

\subsubsection{Compared Methods}
We compare our method with eight state-of-the-art methods, including three SISR methods (\ie Bi-3DQRNN \cite{fu2021bidirectional}, MCNet \cite{li2020mixed}, SSPSR \cite{jiang2020learning}) and five fusion-based methods (\ie, NSSR \cite{dong2016hyperspectral}, Optimized \cite{fu2019hyperspectral}, Integrated \cite{zhou2019integrated}, u2MDN \cite{qu2021unsupervised}, Non-Reg \cite{zheng2021nonregsrnet}). For deep-learning-based methods, we train the networks using the recommended hyperparameters and the same dataset as ours. For optimization-based methods, we empirically select the best parameters to achieve their best performance.

\subsubsection{Evaluation Metrics}

We employ two sets of quantitative quality metrics for systematic evaluation of ours and competing methods. PSNR and SSIM \cite{wang2004image}  are used to evaluate the spatial fidelity of the super-resolved HSI. SAM \cite{yuhas1993determination} is employed to measure spectral similarity. PSNR and SSIM are calculated as the average of the bandwise results for each HSI. Larger values of PSNR and SSIM suggest better performance, while a smaller value of SAM implies better performance.

\begin{table*}
\centering
\setlength{\tabcolsep}{0.27cm}
\caption{Quantitative results on the scale factor of 16 for the simulated and real datasets.}
\label{tab:sf16}
\begin{tabular}{@{}cccccccccccc@{}}
\toprule
 \makecell{Scale\\ Factor}    &   \makecell{Metric\\ ~}   & \makecell{Bicubic \\ ~} & \makecell{Bi-3DQRNN \\ \cite{fu2021bidirectional}}  & \makecell{SSPSR \\ \cite{jiang2020learning}}  & \makecell{MCNet \\ \cite{li2020mixed}}   & \makecell{NSSR \\ \cite{dong2016hyperspectral}}     & \makecell{Optimized \\ \cite{fu2019hyperspectral}}     & \makecell{Integrated \\ \cite{zhou2019integrated}}  & \makecell{NonReg \\ \cite{zheng2021nonregsrnet}} & \makecell{u2MDN \\ \cite{qu2021unsupervised}}      & \makecell{HSIFN \\ (Ours)}  \\ \midrule
\multirow{3}{*}{Simulated} & PSNR $\uparrow$   & 22.10   & 25.34     & 25.31 & 25.42 & 24.23   & 25.39   & 23.57    & 25.53  & 25.31  &\textbf{33.72}       \\
                             & SSIM $\uparrow$   & 0.764   & 0.846     & 0.847 & 0.848 & 0.820   & 0.847   & 0.820  & 0.763  & 0.832  &\textbf{0.963}       \\
                             & SAM $\downarrow$  & 0.151   & 0.100     & 0.104 & 0.106 & 0.136   & 0.318   & 0.370  & 0.360  & 0.151  &\textbf{0.062}       \\ \midrule
\multirow{3}{*}{Real} & PSNR $\uparrow$  & 24.84   & 26.53     & 27.20 & 27.04 & 25.50   & 26.93   & 27.10  & 21.39  & 29.84 &\textbf{31.07}       \\
                     & SSIM $\uparrow$   & 0.815   & 0.832     & 0.851 & 0.845 & 0.918   & 0.912   & 0.885  & 0.123  & 0.931 & \textbf{0.939}       \\
                     & SAM $\downarrow$  & 0.086   & 0.111     & \textbf{0.075} & 0.088 & 0.111   & 0.220   & 0.206 & 0.584 & 0.109 & 0.100       \\ 
                     \bottomrule
\end{tabular}
\end{table*}

\begin{figure*}
 \centering
 \setlength{\tabcolsep}{0.087cm}
  
 \begin{tabular}{cccccc}

  \includegraphics[height=0.072\textwidth]{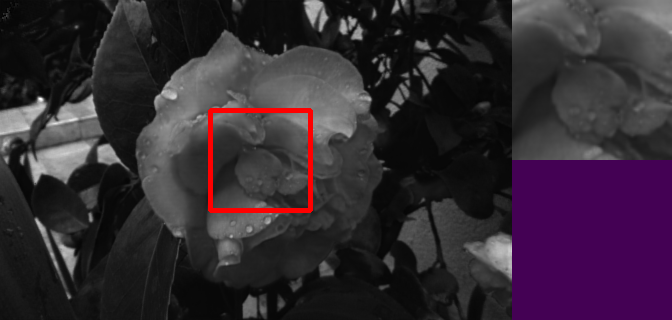}
  & \includegraphics[height=0.072\textwidth]{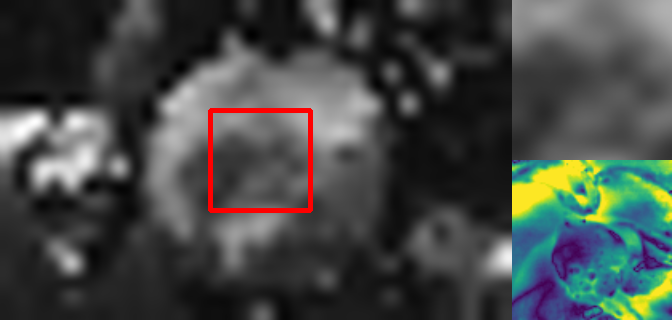}
  & \includegraphics[height=0.072\textwidth]{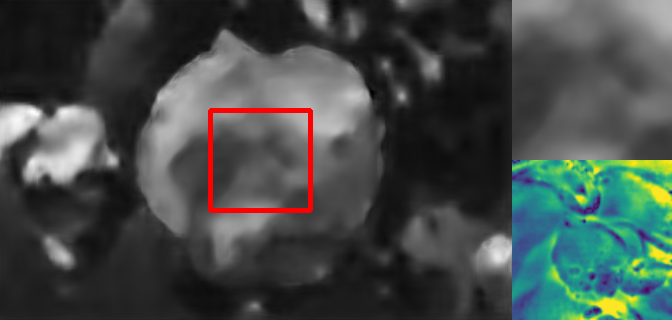}
  & \includegraphics[height=0.072\textwidth]{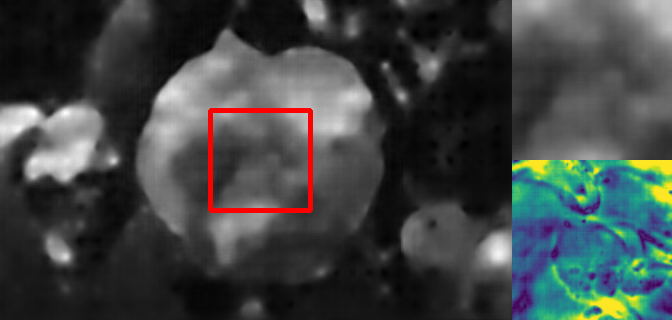}
  & \includegraphics[height=0.072\textwidth]{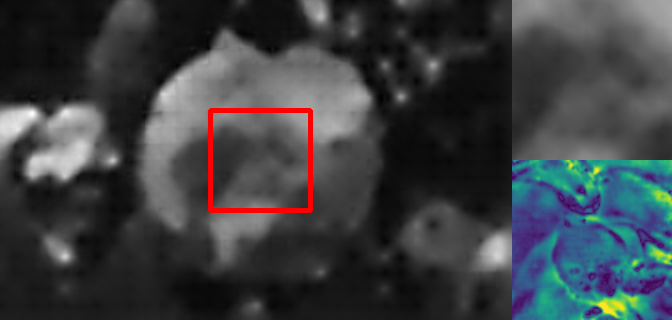}
  & \includegraphics[height=0.072\textwidth]{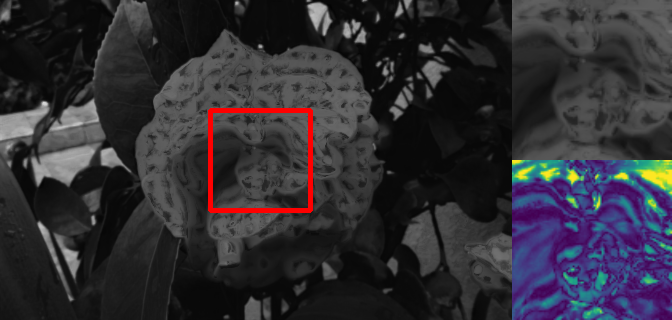}\\
  Ground Truth  & Bicubic & Bi3DQRNN \cite{fu2021bidirectional} & SSPSR \cite{jiang2020learning} & MCNet \cite{li2020mixed} & NSSR \cite{dong2016hyperspectral} \\
  
  \includegraphics[height=0.072\textwidth]{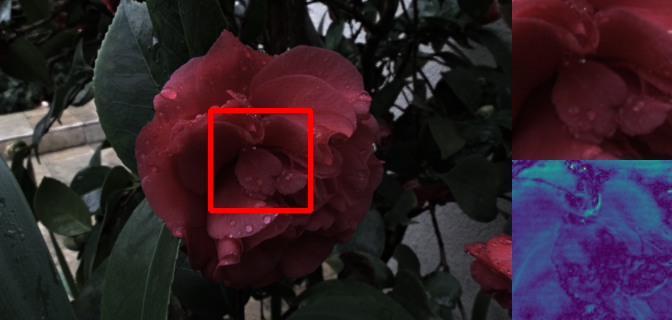}
  & \includegraphics[height=0.072\textwidth]{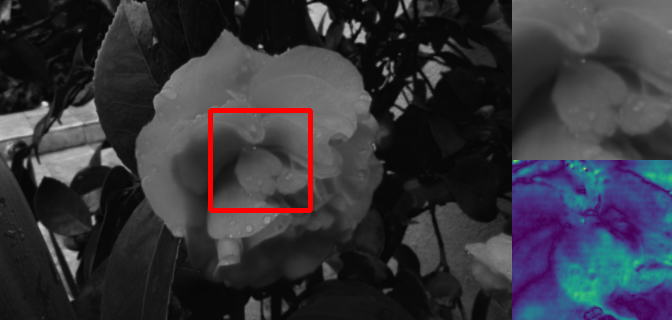}
  & \includegraphics[height=0.072\textwidth]{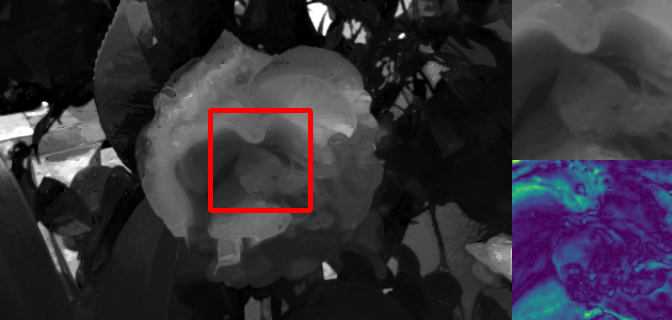}
  & \includegraphics[height=0.072\textwidth]{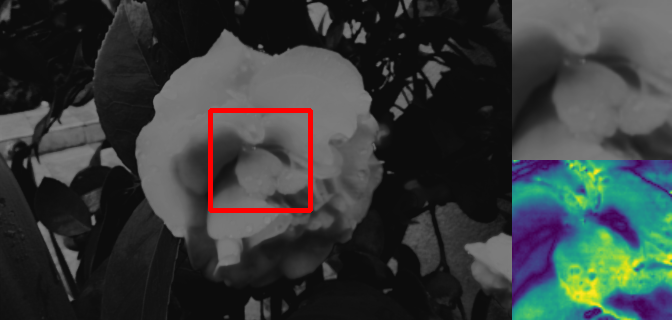}
  & \includegraphics[height=0.072\textwidth]{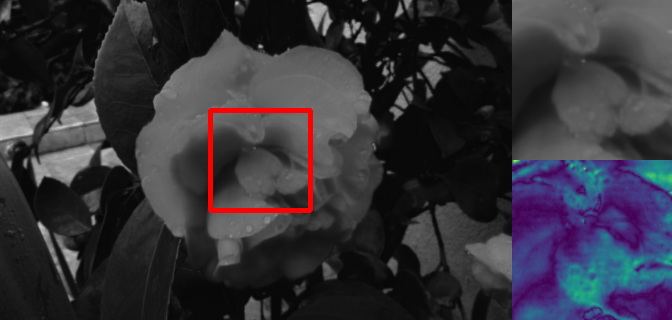}
  & \includegraphics[height=0.072\textwidth]{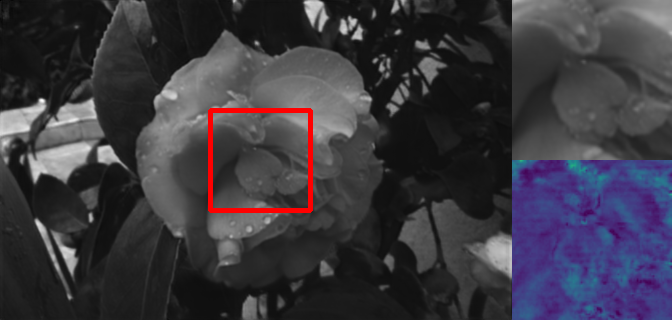}\\
Reference ($\times16$) & Optimized \cite{fu2019hyperspectral} & Integrated \cite{zhou2019integrated} & NonReg \cite{zheng2021nonregsrnet} & u2MDN \cite{qu2021unsupervised} & HSIFN (Ours) \\

  \hline\\ 
  
  \includegraphics[height=0.087\textwidth]{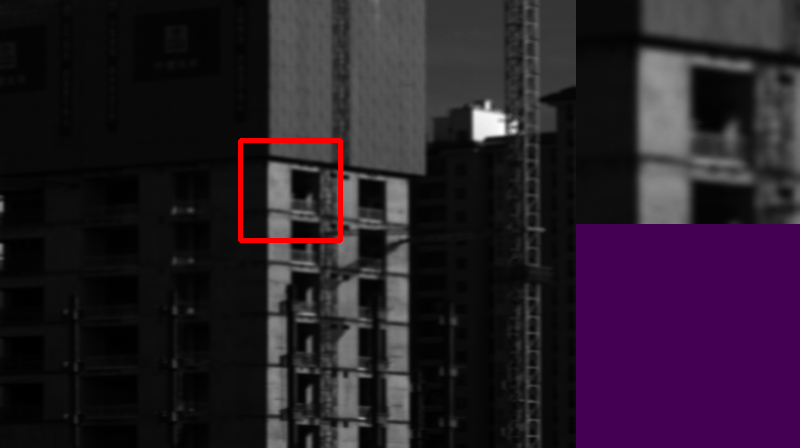}
  & \includegraphics[height=0.087\textwidth]{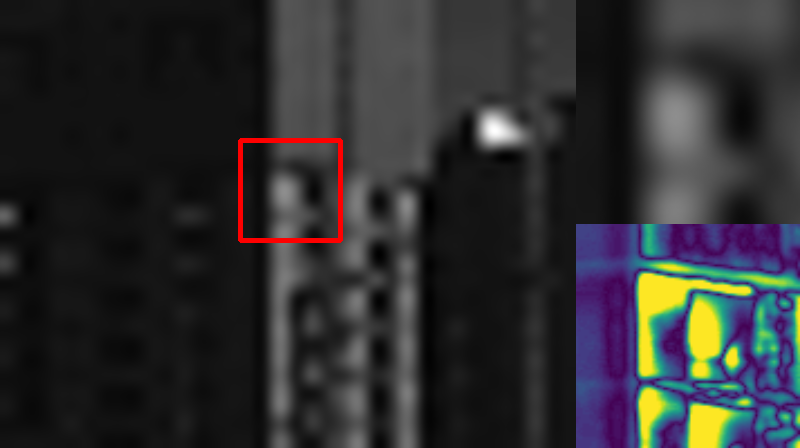}
  & \includegraphics[height=0.087\textwidth]{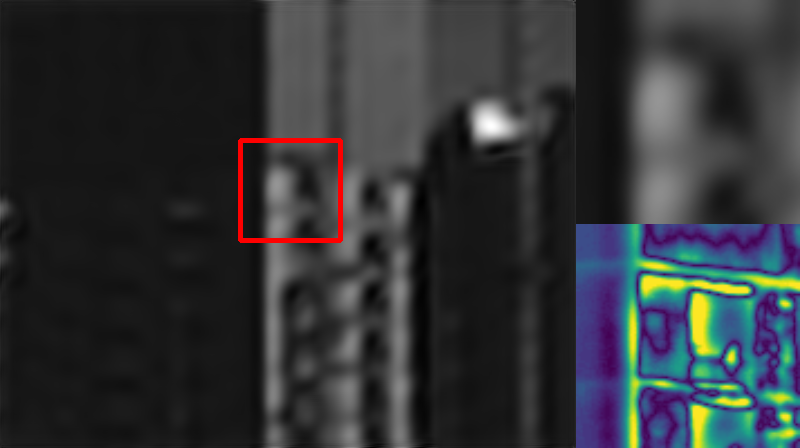}
  & \includegraphics[height=0.087\textwidth]{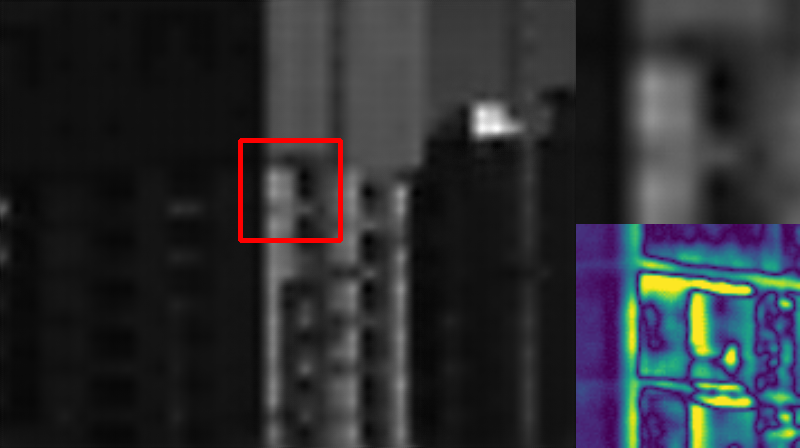}
  & \includegraphics[height=0.087\textwidth]{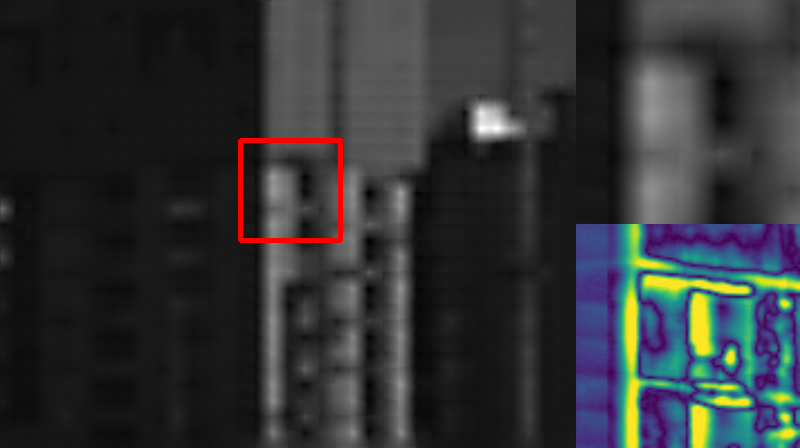}
  & \includegraphics[height=0.087\textwidth]{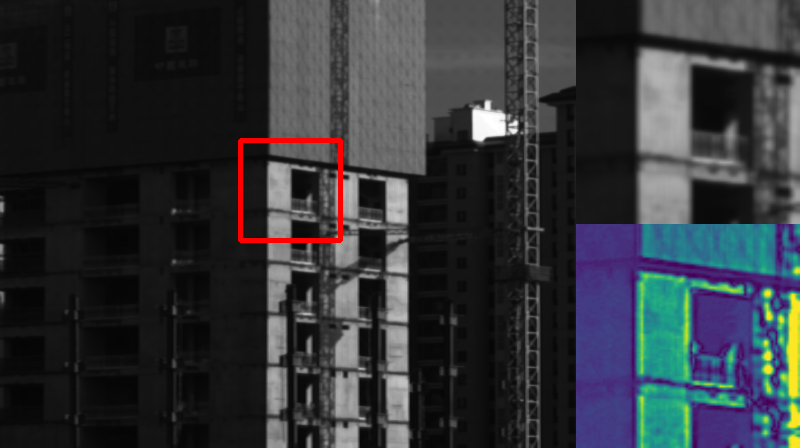}\\
   Ground Truth  & Bicubic & Bi3DQRNN \cite{fu2021bidirectional} & SSPSR \cite{jiang2020learning} & MCNet \cite{li2020mixed} & NSSR \cite{dong2016hyperspectral} \\
  
  \includegraphics[height=0.087\textwidth]{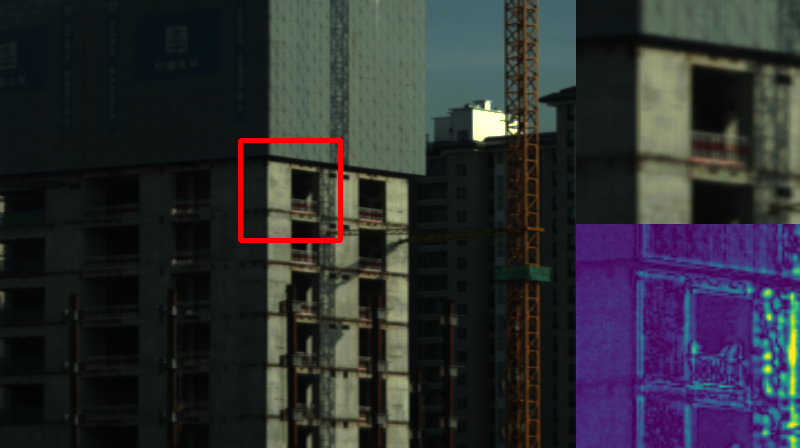}
  & \includegraphics[height=0.087\textwidth]{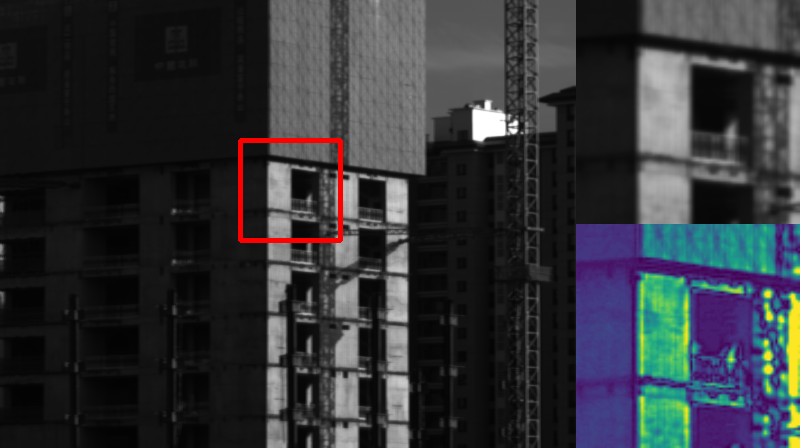}
  & \includegraphics[height=0.087\textwidth]{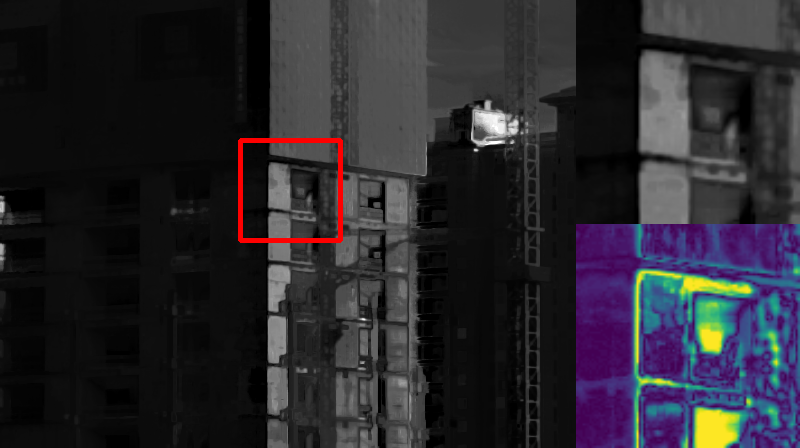}
  & \includegraphics[height=0.087\textwidth]{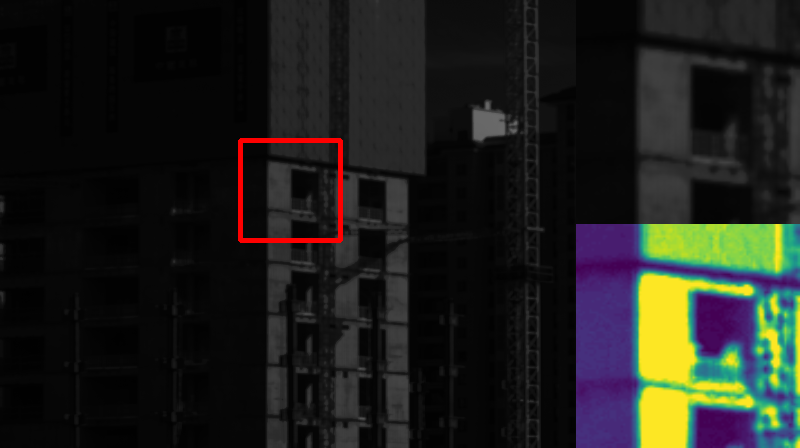}
  & \includegraphics[height=0.087\textwidth]{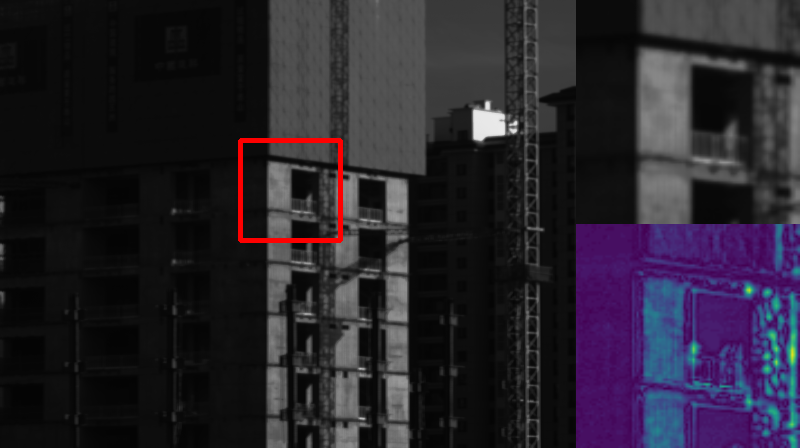}
  & \includegraphics[height=0.087\textwidth]{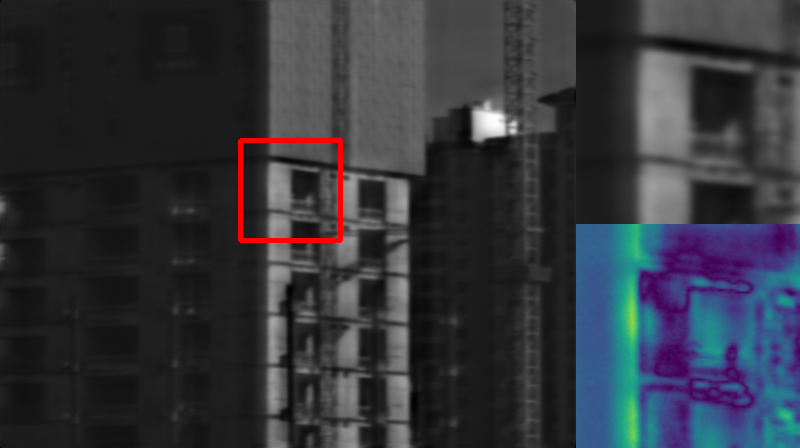}\\
Reference ($\times16$) & Optimized \cite{fu2019hyperspectral} & Integrated \cite{zhou2019integrated} & NonReg \cite{zheng2021nonregsrnet} & u2MDN \cite{qu2021unsupervised} & HSIFN (Ours) \\

 \end{tabular}
 \caption{Visual comparison on the simulated and real dataset under the scale factor of 16. Our method reconstructs more details while properly aligning with the ground truth. Zoom in for details.}
 \label{fig:sf16} 
\end{figure*}

\subsection{Results on Synthetic Data}

In this part, we provide the experimental results on the simulated dataset generated from real unaligned RGB-RGB pairs. The quantitative results are shown in Table \ref{tab:simulate}. It can be observed that our approach is significantly better than all the SISR methods with over 4 and 8 dB improvement on PSNR for scale factors of 4 and 8, respectively. Specifically, the SISR methods could achieve relatively satisfactory results on the scale factor of 4, but their performance significantly drops for a larger scale factor of 8 (over 7 dB on PSNR). This is because SISR relies on context information of LR input to guide the reconstruction of high-frequency details, but such context information is insufficient for higher scale factors, thus resulting in notable performance degradation.
On the contrary, our method retains comparably better performance with a lower decline in PSNR, which demonstrates the advantages of the utilization of additional reference images. 
When compared with other fusion-based methods, our method outperforms them by an even larger margin. Besides, it can be observed that the performance of these methods is similar for different scale factors. The reason for this phenomenon is that even though these fusion-based approaches produce the visually clear outcome as shown in Figure \ref{fig:flower-vis}, their results are, in fact, not properly aligned with the ground truth. On the contrary, our method successfully aligns and transfers the high-frequency details from the reference image and produces the finest results. 
Overall, the experimental results on the simulated dataset indicate that the proper alignment is significantly important for boosting performance on unaligned fusion-based HSI super-resolution.

\subsection{Results on Real Data}

Different from the simulated dataset, the real dataset is notably more challenging due to the larger misalignment and fewer data samples. As the quantitative results are shown in Table \ref{tab:real}, all the competing methods as well as ours suffer from performance drop with respect to PSNR gain over Bicubic when compared with the results on simulated data. Nevertheless, our approach still outperforms all the competing methods in terms of PSNR and SSIM. In particular, three fusion-based methods are notably worse than ours and exhibit similar results on scale factors of 4 and 8. This is largely because the results of these methods are not properly aligned (\ie Integrated \cite{zhou2019integrated}, u2MDN \cite{qu2021unsupervised}, Non-Reg \cite{zheng2021nonregsrnet}), or aligned to the RGB reference (\ie NSSR \cite{dong2016hyperspectral} and Optimized \cite{fu2019hyperspectral}). On the contrary, our model achieves better results by equipping with an alignment module as well as an attention module to align the reference image.  
The visual comparison of different methods is provided in Figure \ref{fig:real-vis}. Our proposed method outperforms other methods in terms of visual quality, generating sharper details while properly aligned to the ground truth. NSSR \cite{dong2016hyperspectral} and Optimized \cite{fu2019hyperspectral} also produce the results with fine details but their results are aligned to the RGB reference image. The other three SISR approaches produce more blurred results than ours as they are unable to utilize the high-frequency information from high-resolution RGB guidance.

\subsection{Ablation Study}

In this section, we provide the results of several ablation studies to investigate our proposed method. We first perform the break-down ablation on the simulated dataset without pre-training to analyze the impact of each components of our network. Then, we verify the effectiveness of each component by comparing it with other variants on our real dataset. 

\begin{table}[]
\centering
\setlength{\tabcolsep}{0.13cm}
\caption{Ablation studies of each component of our network.}
\label{tab:ablation}
\begin{tabular}{@{}ccccccc@{}}
\toprule
Alignment & Attention & Fusion         & Params(M)    & PSNR $\uparrow$   & SSIM $\uparrow$  & SAM $\downarrow$  \\ \midrule
- & - & -         & 1.72   & 31.18 & 0.920 & 0.060 \\
- & - & \checkmark        & 1.97   & 35.82 & 0.971 & 0.050 \\
 \checkmark & -  & \checkmark        &  20.86 & 37.70  & 0.981 & 0.040 \\
\checkmark & \checkmark & \checkmark 		& 21.01 & 38.29  & 0.983 & 0.036 \\ \bottomrule
\end{tabular}
\end{table}

\paragraph{Effect of QRU-based Encoder and Decoder} The quasi-recurrent unit (QRU) is used for merging the features along the spectrum dimension, which is shown to be helpful for improving the quality of the reconstructed HSI. Therefore, we employ the QRU as the basic building block for our HSI encoder and final fusion decoder. To demonstrate the effectiveness of this design choice, we conduct an ablation study that compares the performance of different models w/ and w/o QRU and Bi-QRU. The results are shown in Table \ref{tab:ablation-qru}. Similar to \cite{wei20203}, Bi-QRU is used at the first layer of the HSI encoder, and we alternatively change the direction of the following QRUs to achieve the global contextual receptive field. This strategy provides comparable performance to the full-BiQRU strategy, while reducing the total number of parameters. 

\begin{table}[]
\centering
\setlength{\tabcolsep}{0.1cm}
\caption{Ablation studies of the QRU for encoder and decoder. }
\label{tab:ablation-qru}
\begin{tabular}{@{}llcccc@{}}
\toprule
 Method & Strategy     & Params(M)    & PSNR $\uparrow$   & SSIM $\uparrow$  & SAM $\downarrow$  \\ \midrule
w/o QRU & -     & 20.16   & 40.36 & 0.987 & 0.052 \\
w/ QRU & Alternative     & 20.00   & 41.09 & 0.989 & 0.048 \\
w/ BiQRU & Bi-directional	& 21.85 & 41.22  & 0.989 & 0.048 \\ 
w/ QRU+BiQRU & Alternative       &  21.01 & 41.21  & 0.989 & 0.047 \\
\bottomrule
\end{tabular}
\end{table}

\subsubsection{Effect of Alignment Module} 
\begin{figure}
\centering
\subfigure[LR HSI $\times 8$]{
\begin{minipage}[b]{0.15\textwidth}
\includegraphics[width=1\linewidth]{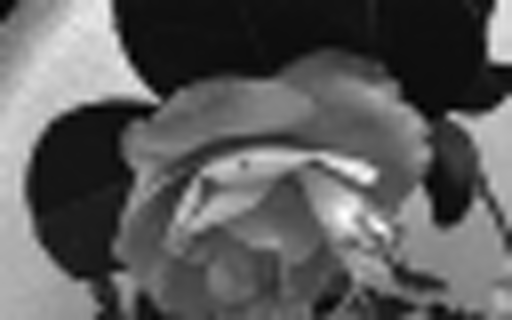}\vspace{2pt}
\includegraphics[width=1\linewidth]{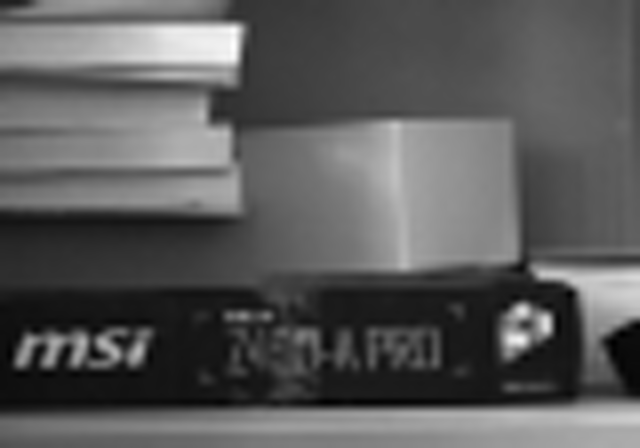}
\end{minipage}}
\subfigure[RGB Reference]{
\begin{minipage}[b]{0.15\textwidth}
\includegraphics[width=1\linewidth]{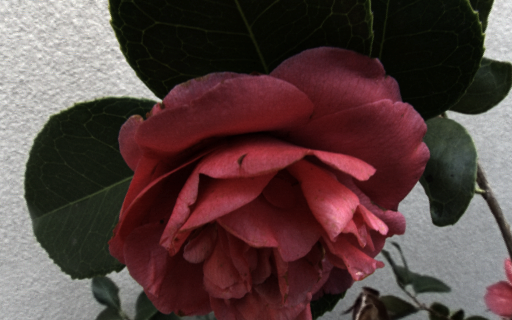}\vspace{2pt}
\includegraphics[width=1\linewidth]{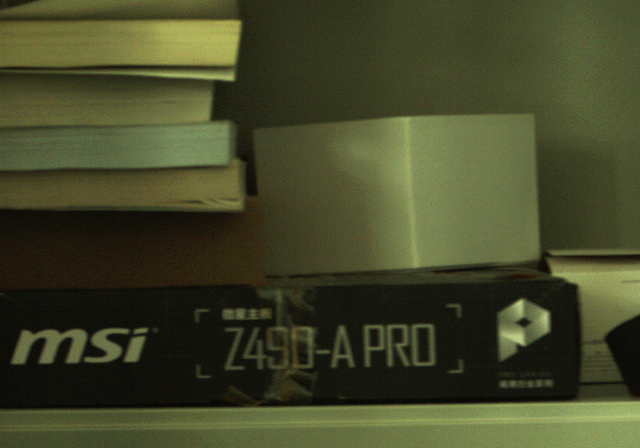}
\end{minipage}}
\subfigure[Predicted Flow]{
\begin{minipage}[b]{0.15\textwidth}
\includegraphics[width=1\linewidth]{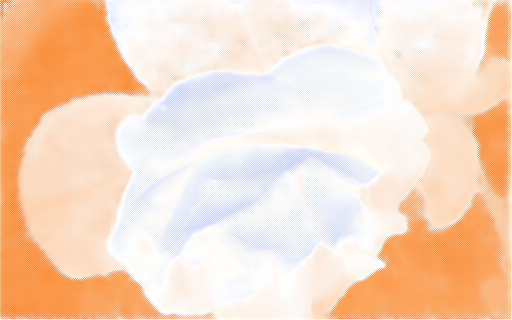}\vspace{2pt}
\includegraphics[width=1\linewidth]{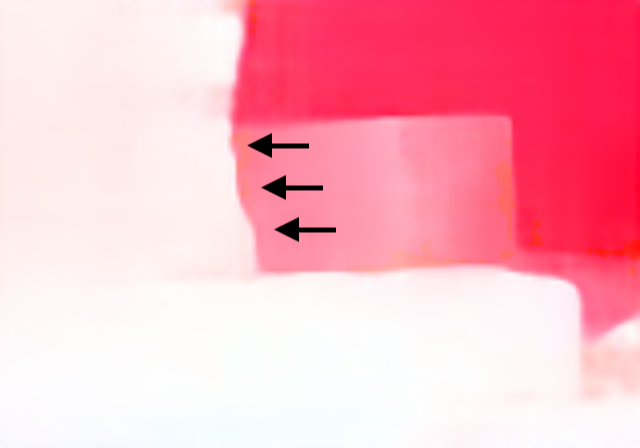}
\end{minipage}}

\caption{Visualization of predicted optical flow between LR HSI and HR reference on simulated (top) and real (bottom) datasets.}

\label{fig:flow-map}
\end{figure}

Alignment is an essential step for the effective fusion of HR RGB reference and LR HSI. As shown in Table \ref{tab:ablation}, the alignment module achieves a significant improvement on PSNR (1.88 dB) and SSIM (0.01) compared with the baseline, which demonstrates its effectiveness. Figure \ref{fig:flow-map} shows the visualization of predicted optical flow on two sample scenes from the simulated and real datasets. We use a pre-trained flow estimator for the real dataset, but the one for the simulated dataset is trained from scratch. Thus, it partially indicates the capability of our model to learn flow estimation even without explicit supervision. 

\begin{figure}
\centering
\subfigure[RGB Reference]{
\includegraphics[width=0.14\textwidth]{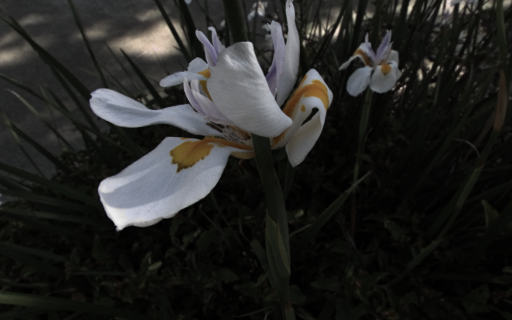}
}
\subfigure[Flow Map]{
\includegraphics[width=0.14\textwidth]{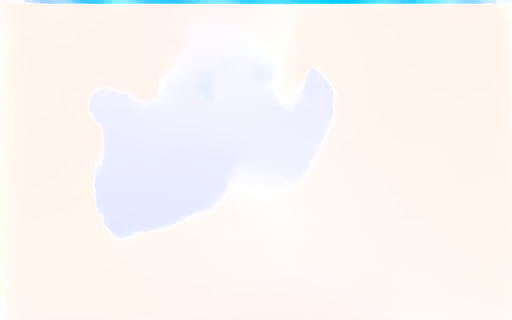}
}
\subfigure[Attention Map]{
\includegraphics[width=0.14\textwidth]{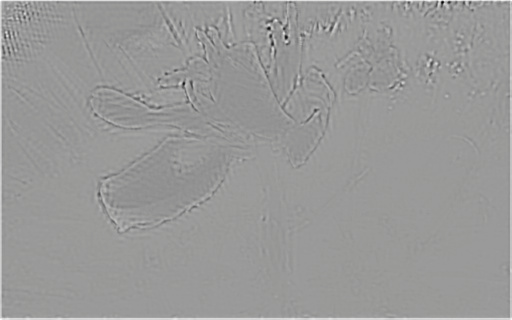}
}
\caption{Visualization of predicted attention map on a sample scene from the simulated dataset.}
\label{fig:attention-map}
\end{figure}

\paragraph{Effect of Attention Module} The ablation results of the attention module are shown in Table \ref{tab:ablation}, we can observe that there is a prominent improvement (0.59 dB on PSNR) after adding the attention module, which verifies its effectiveness. In order to analyze the actual transformation that the attention module learned, we conduct a series of visualizations. As shown in Figure \ref{fig:attention-map}, the generated attention map generally outlines the main objects with some inclinations on the edges and corners.  The reason that the attention module outlines the edges might come from the fact that the displacement of the unaligned image might not produce a difference for non-edges areas. Therefore, the network learns to attend to areas of reference RGB image that differ from the LR HSI, which are more likely to be areas near the edges. Nevertheless, the actual attention map might encapsulate more complex relations beyond the attention on edges.

\begin{table}[]
\centering
\setlength{\tabcolsep}{0.26cm}
\caption{Ablation study of the effectiveness of the pretraining of the optical-flow estimator. }
\label{tab:ablation-flow}
\begin{tabular}{@{}lcccc@{}}
\toprule
Model          & Params (M)    & PSNR $\uparrow$   & SSIM $\uparrow$  & SAM $\downarrow$  \\ \midrule
With Pretrained   &  21.01  & 41.21  & 0.989 & 0.047 \\
Without Pretrained & 21.01  & 36.98  & 0.970 & 0.051 \\ \bottomrule
\end{tabular}
\end{table}

\begin{figure*}[t]
\centering
\subfigure[The spectral curve of some selected pixels. ]{
\includegraphics[width=0.31\textwidth]{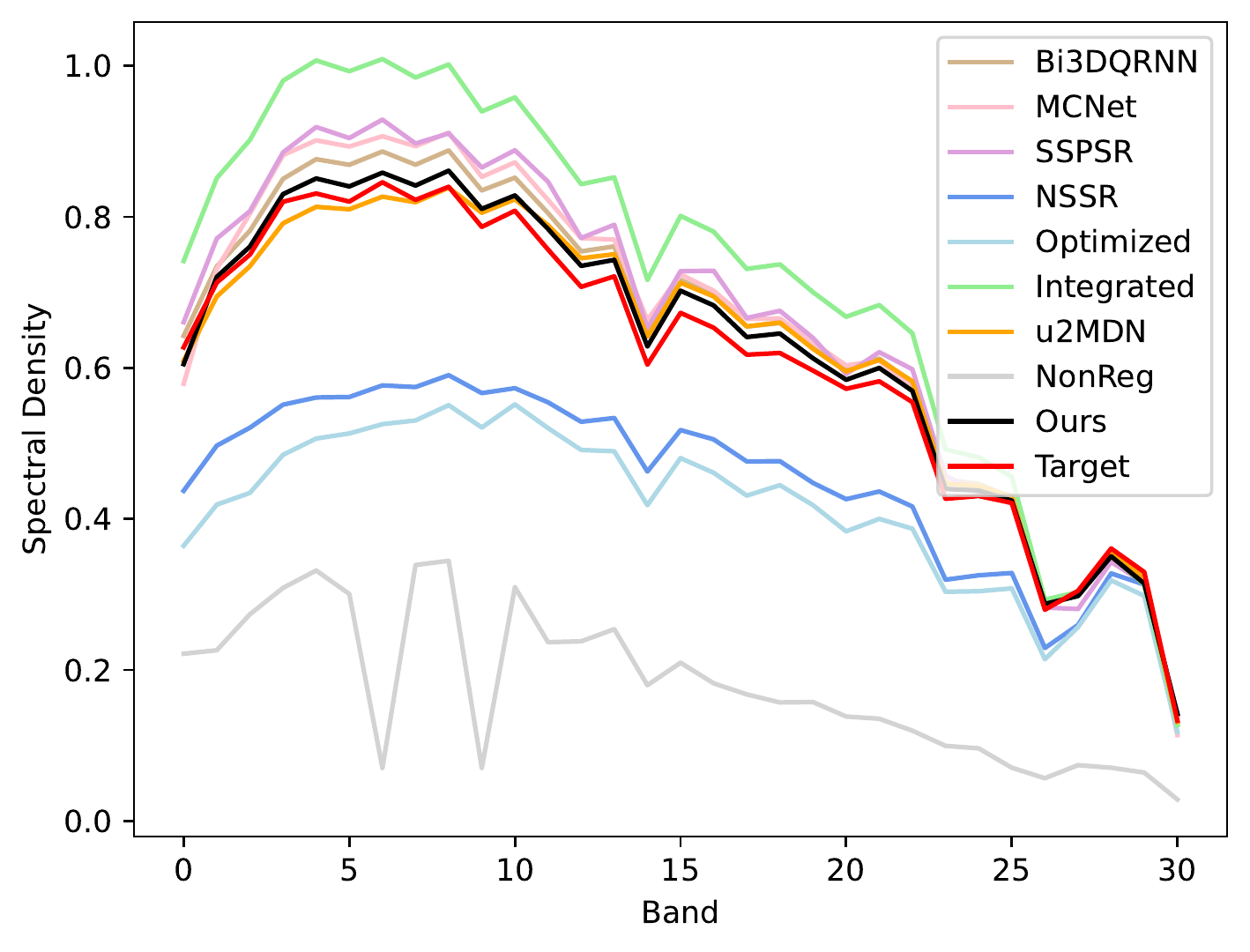}
}
\subfigure[The curve of SSIM for each band.]{
\includegraphics[width=0.31\textwidth]{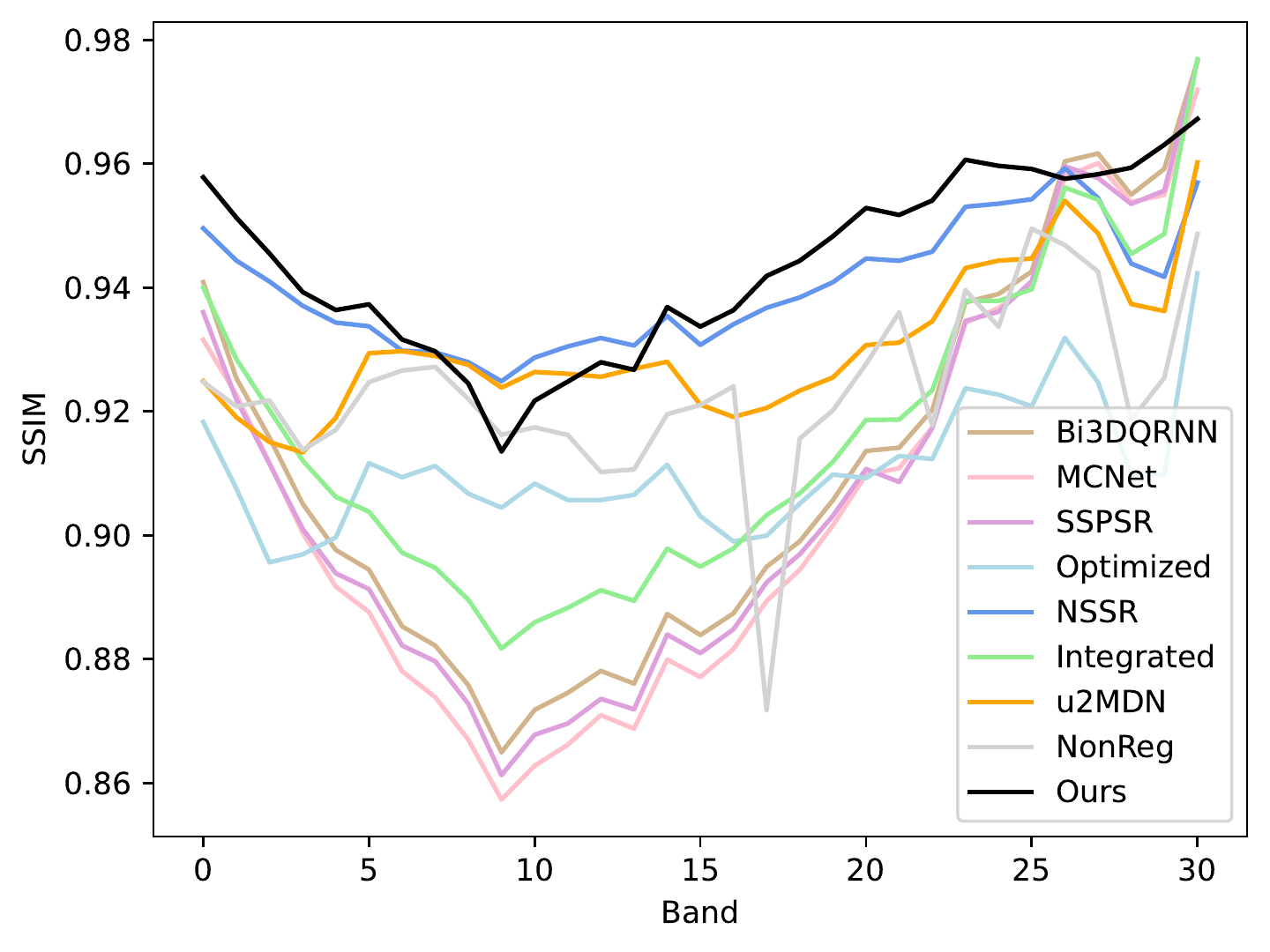}
}
\subfigure[The curve of PSNR for each band.]{
\includegraphics[width=0.31\textwidth]{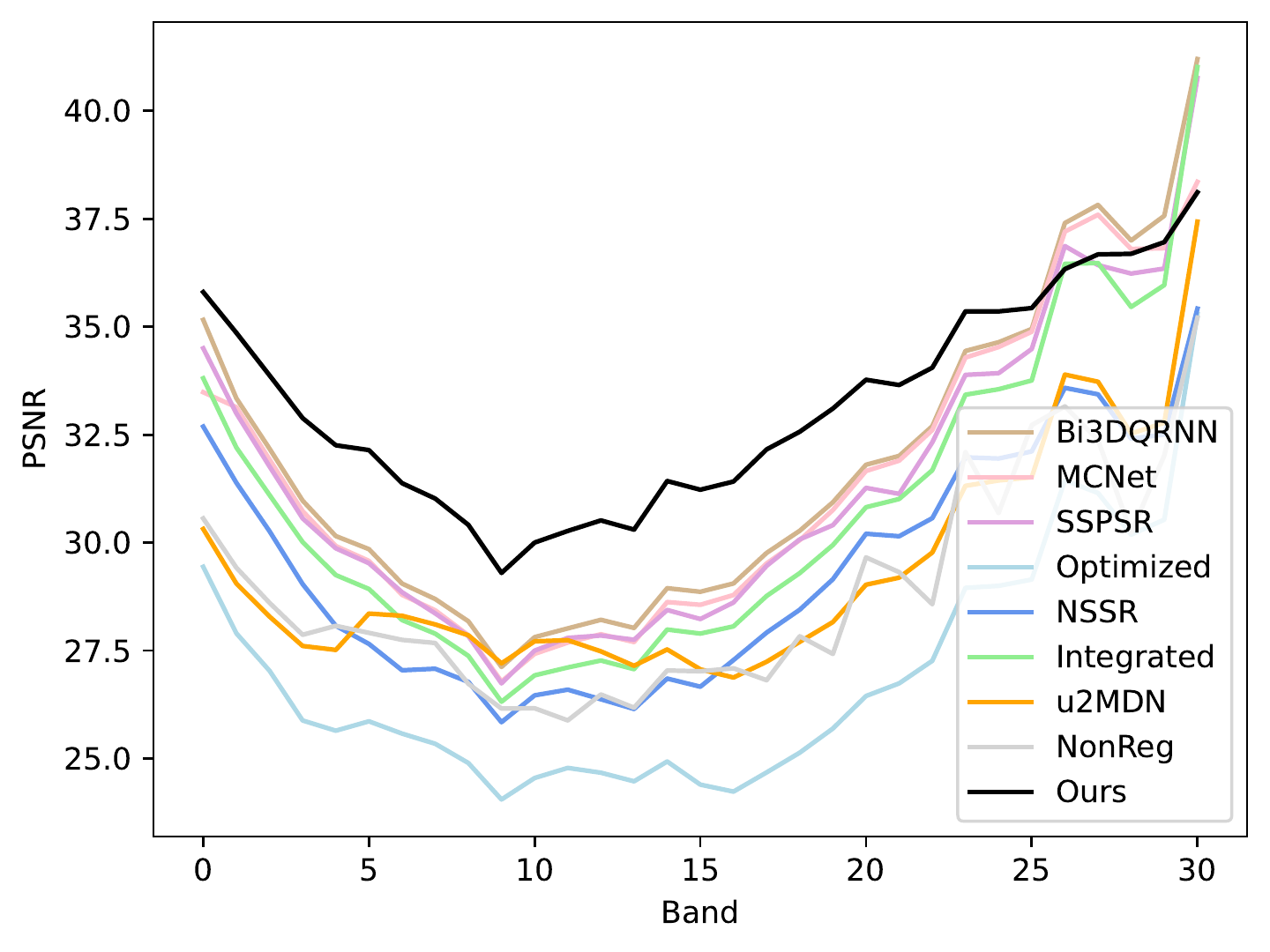}
}
\caption{Visualization of band-wise statistics. (a) Spectral curve: it can be observed that our results (Ours, blue line) is closest to the ground truth (Target, orange line). (b) SSIM curve: it can be seen that our method achieves the best performance in the most spectral bands. (c) PSNR curve: it can be seen that our method achieves the best performance in the most spectral bands.}
\label{fig:bandwise}
\end{figure*}

\paragraph{Effect of Pretrained Flow Estimator}  We also evaluate the effectiveness of the pre-training of flow estimators on the real dataset under the scale factor of 4. The experimental results are shown in Table \ref{tab:ablation-flow}. It can be seen that there is a significant improvement (over 4 dB on PNSR) after adding the pre-training for flow estimators.

\begin{table}[t]
\centering
\setlength{\tabcolsep}{0.23cm}
\caption{Ablation study on the fusion decoder. }
\label{tab:ablation-fusion}
\begin{tabular}{@{}lcccc@{}}
\toprule
Model          & Params (M)    & PSNR $\uparrow$   & SSIM $\uparrow$  & SAM $\downarrow$  \\ \midrule
Decoder without QRU   &  20.45  &  40.56 & 0.988 & 0.051 \\
Decoder with QRU & 21.01  & 41.21  & 0.989 & 0.047 \\ \bottomrule
\end{tabular}
\end{table}

\paragraph{Effect of Fusion Module} We verify the function of the fusion module from two aspects, including (a) The effectiveness of the augmented features from the reference RGB images, and (b) the design of the fusion module, i.e., the use of QRU. 
To verify (a), we construct a variant of our model where the fusion module only takes the LR HSI features as input.  This actually makes our model a SISR model, which also removes the alignment and attention modules. The performance comparison is shown in Table \ref{tab:ablation}. It can be seen that the performance of SISR version severely drops when the augmented reference features are removed. This proves the usefulness of the reference RGB image. 
To verify (b), we remove the QRU in the fusion module and evaluate the model performance. The quantitative results are shown in Table \ref{tab:ablation-fusion}.
It can be seen that the fusion module with QRU is better than the fusion module without QRU. This verifies the effectiveness of the design of our fusion module.

\subsection{Discussion}

\subsubsection{Results on the Large Scale Factor}

Although most of the existing HSI fusion methods consider the significantly larger scale factors, \eg 32, it is physically limited for HSI fusion with unaligned reference to deal with such scale factors. Specifically, suppose there is an HSI in the size of $640\times320$, which is quite common for commercial hyperspectral cameras. After downsampling with scale factor 32, the LR HSI only has 20 pixels in height and 10 pixels in width. However, the fact is that the misalignment of reference and HSI can only be 5 to 10 pixels, which becomes less than one pixel in the downsampled image. This makes it extremely difficult (or even impossible) to align the reference to HSI. Hence, under our task, we choose scale factors 4 and 8 to perform the main experiments. Nevertheless, we further provide the quantitative results on the real and simulated datasets for the scale factor of 16 in Table \ref{tab:sf16}. The visual comparison is shown in Figure \ref{fig:sf16}. It can be seen that our method still outperforms all the competing methods for quantitative results. Overall, for the simulated dataset with sufficient training samples and small misalignment, our method can achieve fairly good performance while obtaining reasonable visual results. For the real dataset, our method may struggle due to difficulty of cross-scale alignment for large misalignment.

\begin{table}
\centering
\setlength{\tabcolsep}{0.38cm}
\caption{The model size of our network with respect to the number of parameters.}
\label{tab:model-size}
\begin{tabular}{|l|l|l|l|}
\hline
HSI Encoder & Attention & FlowNet & Total(FlowNet) \\ \hline
0.58M       & 0.15M     & 33.83$\times$2M       & 69.91M         \\ \hline
RGB Encoder & Decoder   & PWCNet  & Total(PWCNet)  \\ \hline
0.41M       & 1.11M     & 9.37$\times$2M        & 21.01M         \\ \hline
\end{tabular}
\end{table}

\subsubsection{Computational Complexity}

The proposed HSI fusion network contains five different modules, but the overall pipeline is not very complicated and it can be divided into three sequential parts, including feature extraction, feature alignment, and feature fusion. 
(1)	The first part “feature extraction” includes an RGB encoder and an HSI encoder to extract features from reference RGB images and LR HSI. 
(2)	The second part “feature alignment” includes two successive optical-flow estimators and an attention module to align and adjust the features of the reference RGB image. 
(3)	The final part “feature fusion” includes a fusion decoder that fuses the aligned reference features and LR HSI features to predict the final reconstructed HR HSI.
For the quantitative analysis, the total number of parameters of different network components is shown in Table \ref{tab:model-size}. It can be observed that our encoders and fusion decoder are relatively lightweight. The major portion of parameters lies in the optical-flow estimators. Since the design of the optical-flow estimator is not the central topic and contribution of this work, we simply experiment with some classical ones, such as FlowNet \cite{dosovitskiy2015flownet} and PWCNet \cite{sun2018pwc}. However,  our method is not restricted to these flow estimators, and the other estimator, such as RAFT \cite{teed2020raft}, could also be used. For example, the model size can be reduced when RAFT is adopted but the FLOPs would be larger than PWCNet, so it is a trade-off between Params and FLOPs for choosing the flow estimator.

\subsubsection{Bandwise Reconstruction Quality}

To better analysis the reconstruction quality of different methods, we visualize the spectral curve of some selected pixels as well as the curve of SSIM for each band on our real HSI fusion dataset. The visualizations are shown in Figure \ref{fig:bandwise}. For the reconstruction of the spectral curve, it can be seen that our method (black line) is closest to the ground truth. For the curve of SSIM and PSNR for each band, it can be observed that different bands have different performances on the metric of SSIM and PSNR, and our method achieves the best performance on average. The SSIM of SISR methods, i.e., Bi-3DQRNN \cite{fu2021bidirectional}, MCNet \cite{li2020mixed}, and SSPSR \cite{jiang2020learning}, are obviously lower, which is largely due to the absence of the usage of the HR reference image for reconstructing fine-grained details.

\section{Conclusion}
In this paper, we introduce a new unaligned HSI fusion network to address the problem of hyperspectral image super-resolution with real unaligned RGB guidance. 
To deal with the complex misalignment of real unaligned data, a flow-based alignment module is introduced to explicitly align the reference image in the feature space by performing pixel-wise transformation with estimated optical flow. Besides, we propose an element-wise attention module to adaptively adjust the aligned features to drive the network to focus on more discriminative regions, which further improves the performance. 
Moreover, we collect the first HSI fusion dataset with real unaligned pairs of HSI and RGB reference to provide a benchmark and source of training data for unaligned HSI fusion methods. The experiments demonstrate the promising performance and superiority of our unaligned architecture over existing SISR and fusion-based methods. 
We hope our work could provide foundations for further research in the field of HSI fusion with unaligned guidance.


%

%
%

%
%

\ifCLASSOPTIONcaptionsoff
  \newpage
\fi



\bibliographystyle{IEEEtran}
\bibliography{egbib.bib}
%
%
%

%
\begin{IEEEbiography}[{\includegraphics[width=1in,height=1.25in,clip,keepaspectratio]{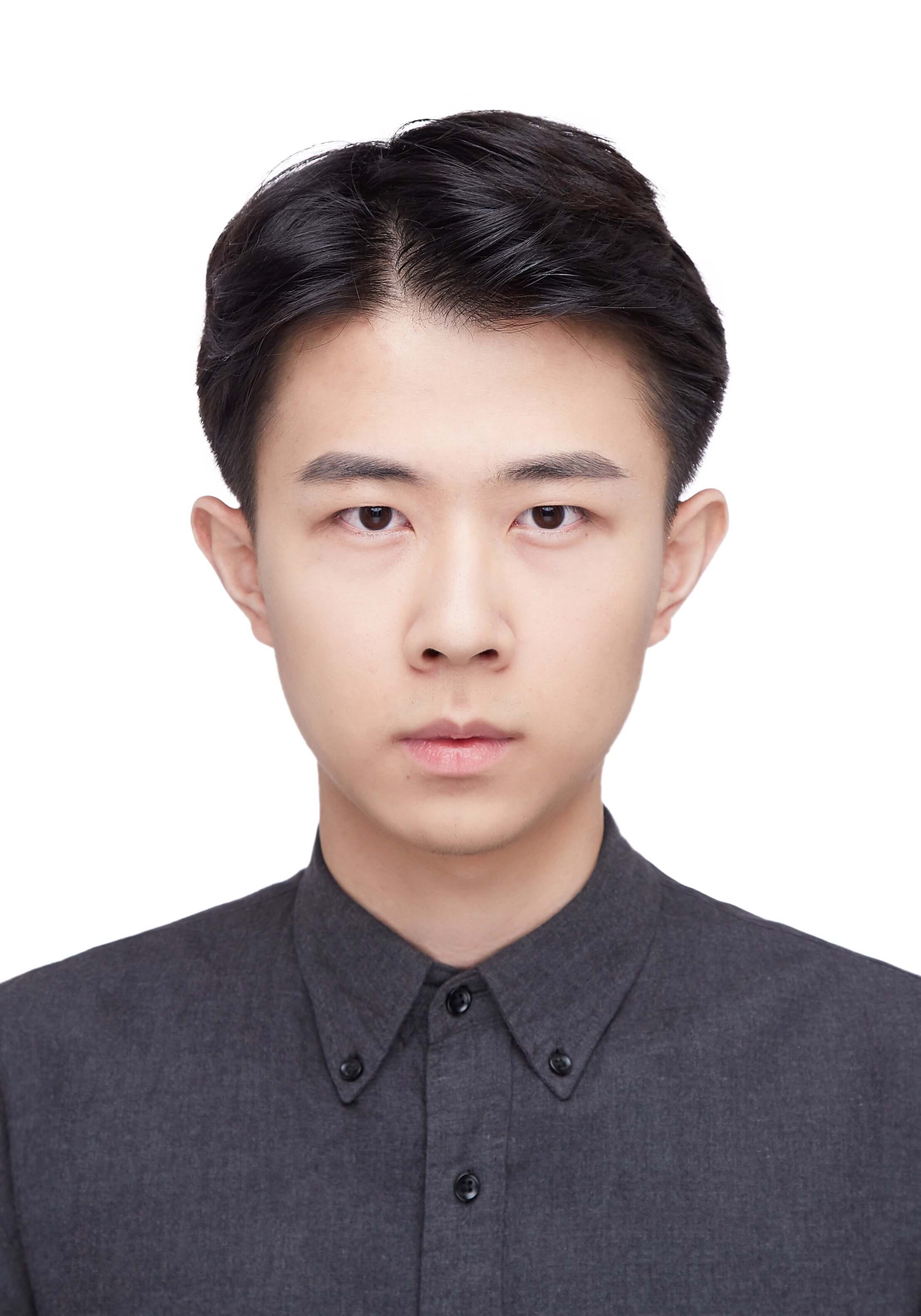}}]{Zeqiang Lai}	
received the bachelor’s degree in computer science and technology from Beijing Institute of Technology, Beijing, China, in 2020. He is currently working toward the master degree from the Beijing Institute of Technology, Beijing, China. His research interests include computer vision, deep learning, and their applications on image processing.
\end{IEEEbiography}

\begin{IEEEbiography}[{\includegraphics[width=1in,height=1.25in,clip,keepaspectratio]{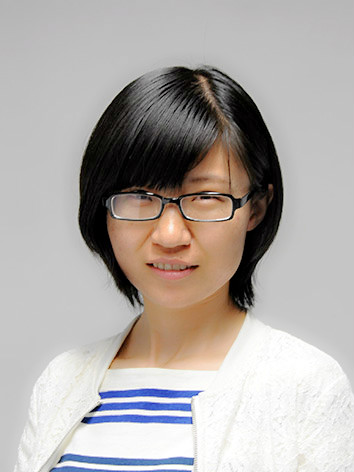}}]{Ying Fu}	
received the B.S. degree in electronic engineering from Xidian University, Xian, China, in 2009, the M.S. degree in automation from Tsinghua University, Beijing, China, in 2012, and the Ph.D. degree in information science and technology from the University of Tokyo, Tokyo, Japan, in 2015. She is currently a Professor with the School of Computer Science and Technology, Beijing Institute of Technology. Her research interests include computer vision, image and video processing, and computational photography.
\end{IEEEbiography}

\begin{IEEEbiography}[{\includegraphics[width=1in,height=1.25in,clip,keepaspectratio]{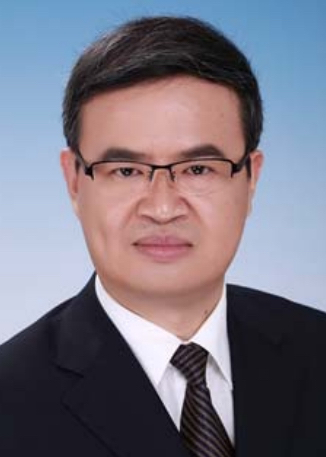}}]{Jun Zhang}	
received the Bachelor, Master and Doctoral degrees in Communications and Electronic Systems from Beihang University in 1987, 1991 and 2001, respectively. He is currently a Professor with Beijing Institute of Technology, where he is also the Secretary for the Party Committee. His research interests are networked and collaborative air traffic management systems, covering signal processing, integrated and heterogeneous networks, and wireless communications. He is a member of the Chinese Academy of Engineering. He has won the awards for science and technology in China many times.
\end{IEEEbiography}









\end{document}